\documentclass[]{article}
\usepackage[a4paper, total={6in, 8in}]{geometry}

\usepackage{booktabs}
\usepackage{graphicx}
\usepackage{multirow}
\usepackage{graphicx,subfigure}
\usepackage{graphicx}
\usepackage{subfigure}
\usepackage{subfloat}
\usepackage{float}
\usepackage{microtype}
\usepackage{authblk}
\usepackage{booktabs}
\usepackage{graphicx}
\usepackage{multirow}
\usepackage{graphicx,subfigure}
\usepackage{graphicx}
\usepackage{subfigure}
\usepackage{subfloat}
\usepackage{float}
\usepackage{microtype}
\usepackage{graphicx,xcolor}
\usepackage{caption}
\usepackage{subcaption}

\usepackage{graphicx}
\usepackage{amsmath}
\usepackage{amssymb}
\usepackage{booktabs}
\usepackage{multirow}
\usepackage{wrapfig}
\usepackage{subfigure}
\usepackage{subcaption}
\usepackage{enumitem}
\usepackage{microtype}
\usepackage[caption=false]{subfig}
\usepackage{url}

\title{Situational Scene Graph for Structured Human-centric Situation Understanding}
\author[1,2,3]{Chinthani Sugandhika}
\author[2,3]{Chen Li}
\author[1]{Deepu Rajan}
\author[2,3,1]{Basura Fernando}
\affil[1]{College of Computing and Data Science, Nanyang Technological University, Singapore.}
\affil[2]{Centre for Frontier AI Research, Agency for Science, Technology and Research, Singapore}
\affil[3]{Institute of High-Performance Computing, Agency for Science, Technology and Research, Singapore}

\date{}

\begin{document}

\maketitle

\begin{abstract}

Graph based representation has been widely used in modelling spatio-temporal relationships in video understanding. Although effective, existing graph-based approaches focus on capturing the human-object relationships while ignoring fine-grained semantic properties of the action components. These semantic properties are crucial for understanding the current situation, such as where does the action takes place, what tools are used and functional properties of the objects. In this work, we propose a graph-based representation called Situational Scene Graph (SSG) to encode both human-object relationships and the corresponding semantic properties. The semantic details are represented as predefined roles and values inspired by situation frame, which is originally designed to represent a single action. Based on our proposed representation, we introduce the task of situational scene graph generation and propose a multi-stage pipeline Interactive and Complementary Network (InComNet) to address the task. Given that the existing datasets are not applicable to the task, we further introduce a SSG dataset whose annotations consist of semantic role-value frames for human, objects and verb predicates of human-object relations. Finally, we demonstrate the effectiveness of our proposed SSG representation by testing on different downstream tasks. Experimental results show that the unified representation can not only benefit predicate classification and semantic role-value classification, but also benefit reasoning tasks on human-centric situation understanding. Code and data are available at \url{https://github.com/LUNAProject22/SSG}.

\end{abstract}    
\section{Introduction}

\begin{figure}[t]
     \centering
     \includegraphics[width=0.6\linewidth]{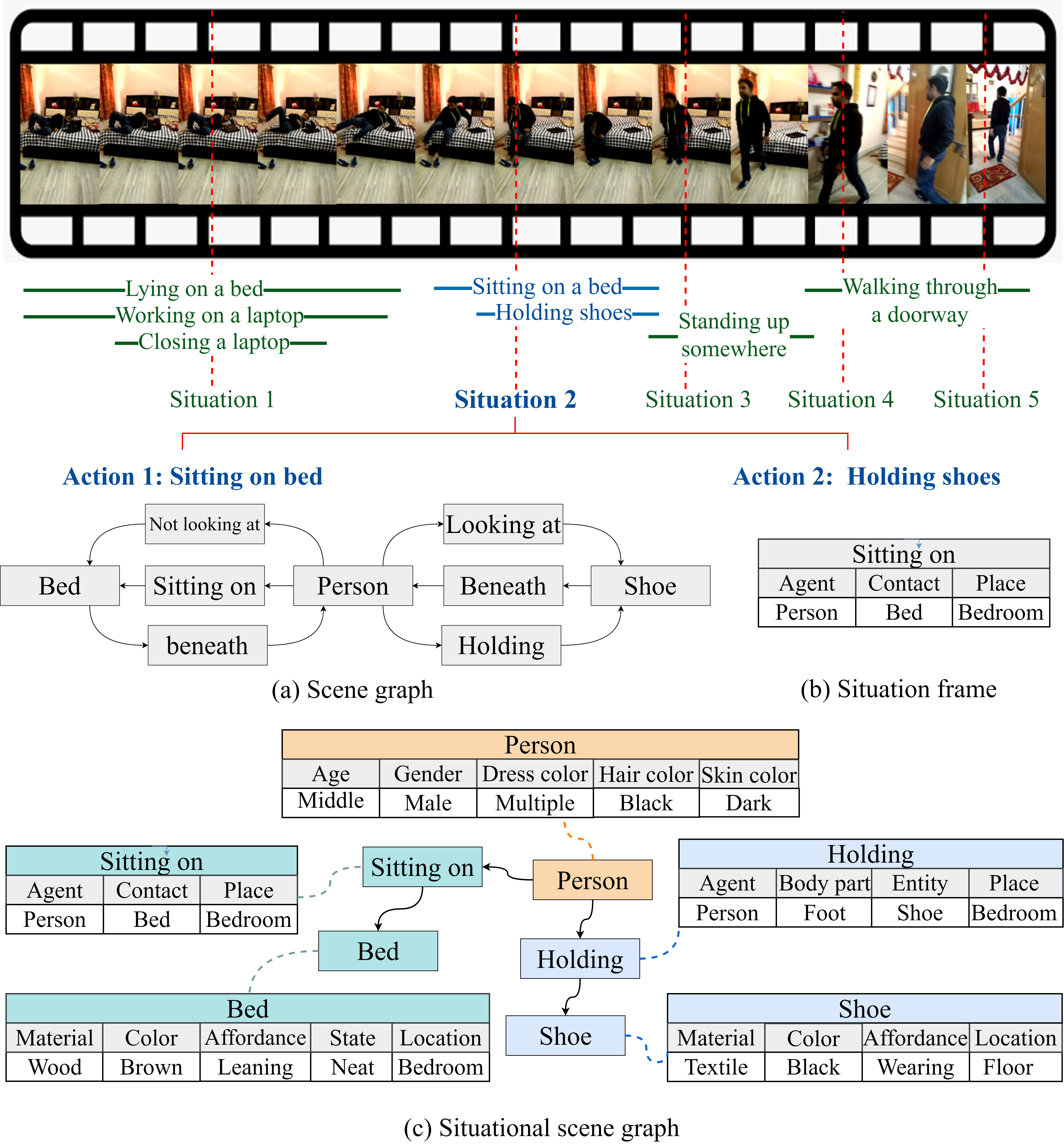}
  \caption{This video frame depicts a human-centric situation of the two concurrent actions "sitting on bed" and "holding shoes". Different structured action representation methods include, (a) Scene graph,
   (b) situation frame, 
   \textbf{(c) Situational scene graph (ours)}: encompasses the person, objects, and verb predicate of human-object relations and their semantic role-values, providing a detailed schema with precisely defined structures to elaborate the components of one or more concurrent actions.
   }
    \label{fig:proposed_representation_method}

\end{figure}

Human-centric visual understanding has traditionally focused on identifying actions occurring in an image or video input \cite{charades, gu2018ava, caba2015activitynet}. 
Although superior performance has been achieved \cite{ryoo2021tokenlearner, kondratyuk2021movinets, wei2022masked}, these studies fail to capture the relationships between human and objects, which may be essential for human behavior understanding. More recently, graph based representations such as spatio-temporal scene graph (ST scene graph) \cite{ag} has been proposed to represent action in a structured manner. The graph based representation encodes human and objects as nodes connected together by pairwise relationships (as shown in Fig. \ref{fig:proposed_representation_method}a), and can support high-level inference tasks such as visual question answering (VQA) \cite{qian2024nuscenes, de2023selfgraphvqa, star, hildebrandt2020scene, johnson2017clevr, teney2017graph, grunde2021agqa}, 
image captioning \cite{yang2019auto, yang2019auto},
referring expressions \cite{yang2019cross}
image grounded dialog \cite{das2017visual} and
image retrieval \cite{johnson2015image} etc. However, the ST scene graph falls short of capturing the fine-grained properties of the relationship components, such as where the action takes place, functional properties of the objects (e.g. affordance) or any tools used by the person to perform the actions etc.
Such fine-grain
properties of human, objects and their pair-wise relationships are
crucial in real-world applications, such as in robotics where the robot needs to understand not only the human behavior but also the properties of the objects (e.g. affordance) in order to interact with them. 

Another line of research \cite{yatskar2016situation, yatskar2017commonly} based on situation frames offer a well-structured schema to represent these semantic properties in terms of predefined
roles and values, providing richer context (as shown in Fig. \ref{fig:proposed_representation_method}b). However, they
are limited to representing a single action, which is not applicable to scenarios with multiple simultaneous actions. In view of the limitations of existing approaches, we propose \textbf{Situational Scene Graph (SSG)} to represent both human-object relations and the corresponding semantic properties in a unified representation.  As shown
in Fig. \ref{fig:proposed_representation_method}c, we leverage the well-structured semantic
role-value (SRV) pairs to encode the properties of the components within verb relationships (i.e. person, object and verb predicate) of a spatio-temporal scene
graph. Our SSG is beneficial for \textbf{Human-centric situation understanding} where we aim to understand
the actions, 
human-object relations as well as semantic properties of each entity involved in the current situation. Moreover, another advantage of SSG is its ability to jointly model the human-object relations and the semantic properties which could benefit both semantic role-value prediction
and predicate prediction.
Take the scene of a person holding a cup as an example. The relationship between the person and the cup (i.e., `holding') 
implies that the person is likely engaging in an action related to the cup such as drinking. Additionally, 
knowing the semantic properties of the cup (e.g., affordance) 
narrows down the possible actions or relations, 
and the relationship `holding' also provides cues for the semantic properties of the cup.
This mutual reinforcement enhances the performance of both semantic role-value prediction and predicate prediction.

Based on our proposed representation, we introduce the task of SSG generation and introduce a new framework named \textbf{In}teractive and \textbf{Com}plementary \textbf{Net}work
(InComNet).  Our InComNet can be factorized into four stages: (I) object SRV classification (II)
verb predicate classification (III) verb predicate SRV classification and (IV) person SRV classification. Stage
(II) classifies the verb predicates while stage (I), (III)
and (IV) classify the semantic role-value of each semantic
role of object, verb predicate of the relation instances and
person respectively. Given that existing datasets are not applicable to this
task, we further introduce a SSG dataset based on the Action Genome \cite{ag} annotations by manually collecting 562K+ situation frames. 

In summary, our key contributions can be listed as follows. Firstly, we introduce a unified SSG representation to encode both human-object relations and the corresponding semantic properties. Secondly, we introduce the challenging task of SSG generation and propose the InComNet baseline. Thirdly, we introduce the SSG dataset, which consists of
2.5K video clips with human-generated annotations for 25.5K human-centric situations encompassing \textbf{25.5K} person SRV frames, \textbf{61K} object SRV frames, \textbf{52K} verb predicate SRV frames.
Lastly, we demonstrate the effectiveness of our proposed SSG representation by testing on different downstream tasks. Experimental results show that the unified representation can not only benefit semantic role-value and predicate prediction, but also benefit reasoning tasks on human-centric situation understanding.

\section{Previous work}
\label{sec:Previous Work}

\sloppy
\noindent
\textbf{Structured visual representation methods:}
Classical structured scene representation methods include scene graphs \cite{johnson2015image} and situation frames \cite{yatskar2016situation}. 
Recent advancements such as spatio-temporal scene graphs \cite{ag} and spatio-temporal situations \cite{sadhu2021visual}
leveraged the principles laid down by these classical methods to enhance video understanding. 
Other methods that build on them include STAR \cite{star}, scene graphs fusion \cite{wu2021scenegraphfusion}, panoptic scene graph \cite{yang2022panoptic} and Panoptic Video Scene Graph (PVSG) \cite{yang2023panoptic}. Some notable advancements brought by the above methods include, improving the pixel level accuracy in localization \cite{yang2022panoptic, yang2023panoptic} and sub-scene graph fusion \cite{wu2021scenegraphfusion}. 
In contrast, situational scene graph takes a unique approach by seeking to leverage the strengths of both scene graph and situation frame to construct a more detailed schema with properly defined structures for representing and encoding components within action while also enabling to capture situations with multiple concurrent actions.
Prior work have introduced problem tasks paired with the above representations such as, spatio-temporal/panoptioc scene graph generation \cite{ag, yang2022panoptic}, situation recognition \cite{yatskar2016situation} and video semantic role-labelling \cite{sadhu2021visual}. Similarly, based on our representation we also introduce a challenging problem task called situational scene graph generation.

\sloppy
\noindent
\textbf{Existing datatsets:}
Table \ref{tab:datasets} provides some popular video understanding benchmarks that are related to us. One major trend in the early stage is the provision of a large number of video clips with single action labels. Around year 2020, the community started focusing on action decomposition to delve into dynamics within actions. 
Although these decompositions offer somewhat holistic schemas for representing actions—such as person, object, and relationships, they often fall short in providing structured methods for encoding the properties of those decomposed elements (refer to Section \ref{sec:ssg_dataset} for a comparison of existing action representations and their required annotations).
Following this trajectory, we introduce the SSG benchmark, which further decomposes these action elements into their sub-semantic properties by utilizing semantic roles and values. The benchmark includes a total of \textbf{562K} human-generated semantic role-value annotations.

\begin{table*}[t!]

  \centering
  \resizebox{1\textwidth}{!}{
  \begin{tabular}{|@{}l|c|c|c|c|c|c|c|c|c|c|c|c|c|c@{}|}
    \hline
    \multirow{2}{*}{Dataset} & \multirow{2}{*}{\# videos} & \multirow{2}{*}{\# hours} &  \multirow{2}{*}
    {\# actions}
    & \multicolumn{2}{c|}{Objects} &  \multicolumn{2}{c|}{Verb predicates} & \multirow{2}{*}{\# persons} &  \multicolumn{2}{c|}{Object SR} & \multicolumn{2}{c|}{Verb predicate SR} & \multicolumn{2}{c|}{Person SR} \\

    \cline{5-8} \cline{10-15}

    & & & & \multicolumn{1}{c|}{\# categories} & \multicolumn{1}{c|}{\# instances} &  \multicolumn{1}{c|}{\# categories} & \multicolumn{1}{c|}{\# instances} & & \multicolumn{1}{c|}{\# roles} & \multicolumn{1}{c|}{\# instances} & \multicolumn{1}{c|}{\# roles} & \multicolumn{1}{c|}{\# instances} & \multicolumn{1}{c|}{\# roles} & \multicolumn{1}{c|}{\# instances} \\
    
    \hline

        ActivityNet (2015) \cite{caba2015activitynet} & 28K & 648 & 200 & - & - & - & - & - & - & - & - & - & - & - \\

        DALY (2016) \cite{weinzaepfel2016human} & 8k & 31 & 10 & 41 & 3.6K & - & - & - & - & - & - & - & - & - \\

        Charades (2016) \cite{charades} & 10K & 82 & 157 & 37 & - & - & - & - & - & - & - & - & - & - \\

        AVA (2018) \cite{gu2018ava}& 504K & 108 & 80 & - & - & 49 & - & - & - & - & - & - & - & - \\

        EPIC-Kitchen (2018) \cite{damen2018scaling} & - & 55 & 125 & 331 & - & - & - & - & - & - & - & - & - & - \\
        
        HACS Clips (2019) \cite{zhao2019hacs} & 0.4K & 833 & 200 & - & - & - & - & - & - & - & - & - & - & - \\
        
        Kinetics-700 (2019)\cite{carreira2019short} & 650K & 1794 & 700 & - & - & - & - & - & - & - & - & - & - & - \\

        CAD120++ (2019) \cite{zhuo2019explainable} & 0.5K & 0.57 & 10 & 13 & 64K & 6 & 32K & - & - & - & - & - & - & - \\
        
        Action Genome (2020) \cite{ag} & 10k & 82 & 157 & 35 & 0.4M & 25 & 1.7M & 220K & - & - & - & - & - & - \\
        
        VidSitu (2021) \cite{sadhu2021visual} & 29.2K & 83 & - & 5.6K & - & 1.5K & 397K & - & - & - & 5 & \textbf{202K} & - & - \\
        
        STAR (2021) \cite{star} & 22K & 80 & - & 37 & - & 24 & - & - & - & - & - & - & - & - \\
        
        PVSG (2023) \cite{yang2023panoptic} & 0.4K & 9 & - & 257 & 7.5K & 57 & 4.1K & 1.1K & - & - & - & - & - & - \\
        \hline
        SSG (ours) & 2.5K & 22 & 157 & 35 & 61K & 16 & 52K & 25.5K & \textbf{16} & \textbf{271K} & \textbf{12} & 164K & \textbf{5} & \textbf{127K} \\
        
  \hline
  \end{tabular}%
  }
  \caption{An analysis of SSG in comparison to other publicly available video understanding benchmarks. SR refers to semantic roles.}
  \label{tab:datasets}
\end{table*}

\noindent\textbf{Visual representational models:}   
To generate scene graphs and situation frames from images, the approaches explored include Conditional Random Field models \cite{johnson2015image, yatskar2016situation, yatskar2017commonly}, Recurrent Neural Networks \cite{xu2017scene, mallya2017recurrent, pratt2020grounded}, Graph Neural Networks \cite{yang2018graph, li2021pose, qian2019video, li2017situation}, few-shot learning \cite{chen2019scene} and transformers \cite{cong2023reltr, cho2021grounded, wei2022rethinking, cho2022collaborative}. For video data, transformers have been widely used to generate the spatio-temporal scene graphs and spatio-temporal situation frames ~\cite{sttran, li2022dynamic, lu2021context, yang2023panoptic, teng2021target, sadhu2021visual}. 
Meanwhile, Large Vision Language Models (LVLMs) like CLIP \cite{clip}, ALIGN \cite{jia2021scaling}, LLaVa \cite{liu2024improved} and VILA \cite{lin2024vila} have been instrumental in developing powerful joint semantic representations of vision and language. 
Different vision tasks
try to leverage these LVLMs in various ways. For example,
recent methods have employed the CLIP embeddings for situation recognition \cite{clipsitu} and action recognition \cite{jiang2023exploiting} attaining excellent results in respective problems.
Inspired by this line of research, we also utilize CLIP embedding to solve our new problem situational scene graph generation.%

\section{Situational Scene Graph}
\label{sec:Method}
 
\begin{figure}[t]
    \centering
    \includegraphics[width=0.3\linewidth]{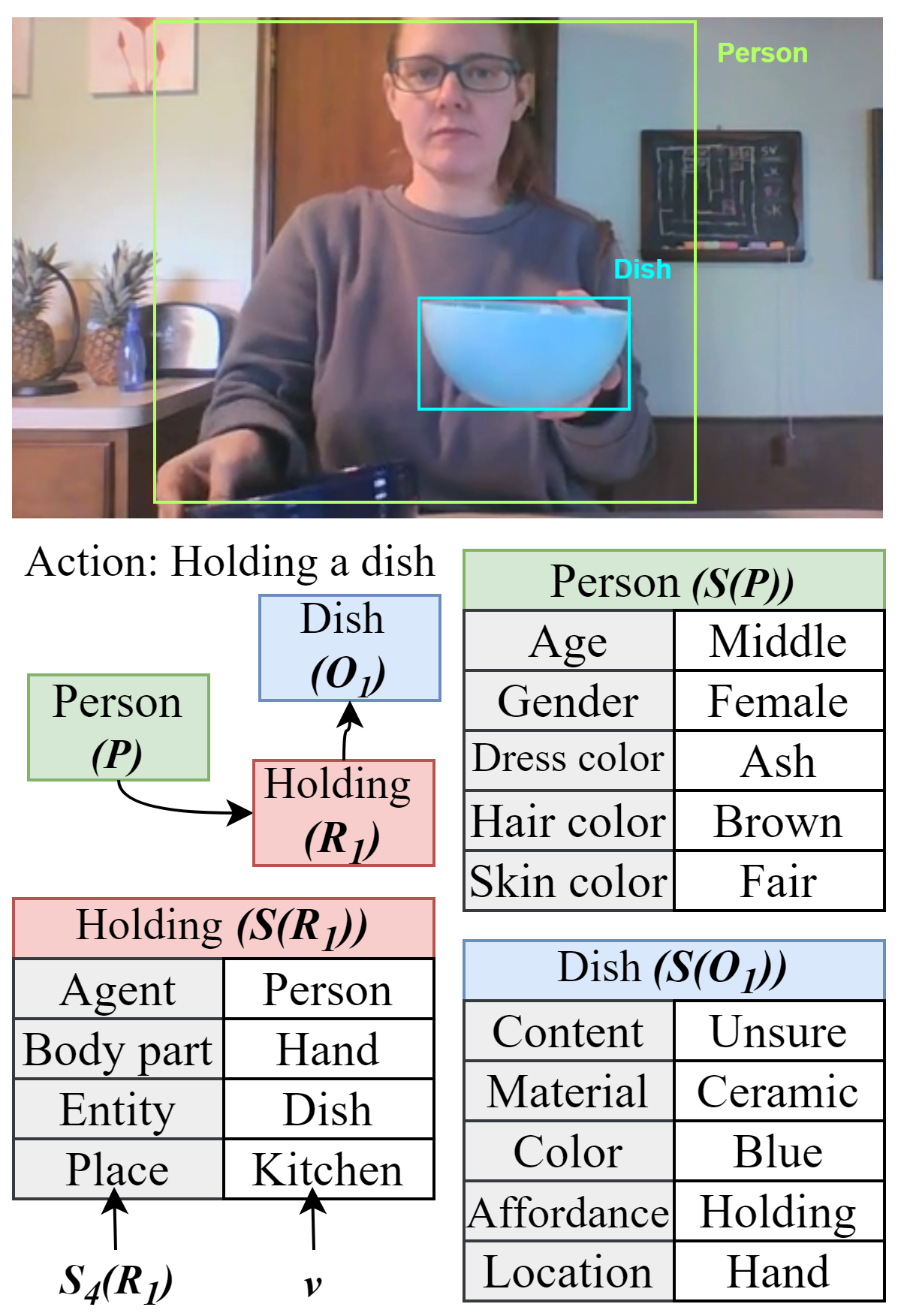}
  \caption{This video frame illustrates a situation of the action 'holding a dish'. Tables, employing color codes green, blue and red depict the semantic roles and their associated values for the semantic entities person, object and verb predicate of the relation instance. }
    \label{fig:method}
\end{figure}

A Situational Scene Graph (SSG) is formed by a set of semantic role-value pairs that elaborates several important semantic entities in a situation(s). 
As shown in Fig. \ref{fig:method}, these semantic entities include the actor or the person, the object instances the person interacts with, and the verb predicate of human-object relation instances. Furthermore, the overall situation may comprise the execution of one or more high-level human actions.

\sloppy
Formally, let us define the set of object classes as $\mathcal{O} = \{O_1, \cdots, O_n \}$, and the person category by $P$. 
Person instance in a video frame forms relations with object instances where each relation is represented by a triplet of the form $\left< \emph{Person, Predicate, Object} \right>$, e.g., \textlangle person, holding, dish\textrangle~in Fig. \ref{fig:method}.
In the SSG, verb predicates like "holding," "drinking from," and "sitting on" are used to describe human-object interactions because they provide more detailed information about actions than spatial predicates like "in," "under," or "beneath"~\cite{goel2022not}.
The set of verb predicate classes is denoted by $\mathcal{R} = \{R_1, \cdots, R_m \}$.
The novelty in the SSG constructs lies in the fact that we annotate 
the semantic role-value pairs for each entity category and instance (person, object and verb predicates) in a scene graph of a person performing one or more high-level actions. 
The semantic role-value pair structure, known as the \emph{Frame Structure}, is predefined based on the literature~\cite{yatskar2016situation,pratt2020grounded}.
The set of semantic roles for person class is predefined and is denoted by $S(P) = \{S_1(P), S_2(P), \cdots \}$.
Similarly, the semantic roles of each object category are also predefined and denoted by $S(O_k) = \{S_1(O_k), S_2(O_k), S_3(O_k), \cdots \}$ for each object category $O_k$.
Finally, the semantic roles of each verb predicate category are denoted by $S(R_k) = \{S_1(R_k), S_2(R_k), S_3(R_k), \cdots \}$ for each verb predicate category $R_k$.
The associated values $v$ of each semantic role (i.e. $S(P), S(O), S(R)$) depend on the specifics of the person, object and verb predicate of the relation instances in a given frame (e.g. $S_{Place}(Holding) = Kitchen$).
The set of all distinct semantic-role values is denoted by $\mathcal{V} = \{v_1, v_2, \cdots \}$. 
In each video frame, there is 
one person instance executing one or more high-level human actions forming relations with object instances.
Each person, object, or relation instance's semantic role is associated with a value
\begin{equation}
    S(A) = v.
    \label{eq.value.classification}
\end{equation}%
where $A \in \{P,\mathcal{O},\mathcal{R} \}$ and $v \in \mathcal{V}$.

\begin{figure}[t]
   \centering
     \includegraphics[width=1\linewidth]{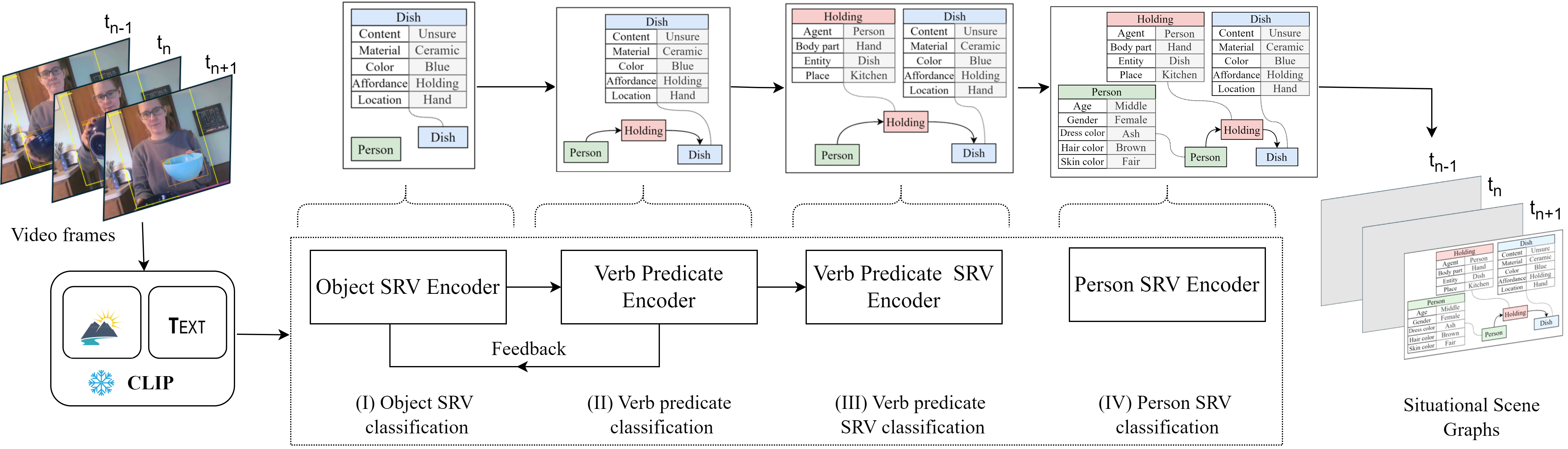}
     \caption{The pipeline of our proposed InComNet. Given a set of video frames, our model uses CLIP to extract necessary feature embeddings from each frame and then classifies SRV of objects, verb predicates, SRV of verb predicates and SRV of person. Finally, the situational scene graph is obtained on the right side. The InComNet stage (II) correspond to the SSG sub-task (1) and stages (I), (III) and (IV) correspond to the SSG sub-task (2).}
    \label{fig:architecture}

\end{figure}

\begin{figure}
    \centering
    \includegraphics[width=0.3\textwidth]{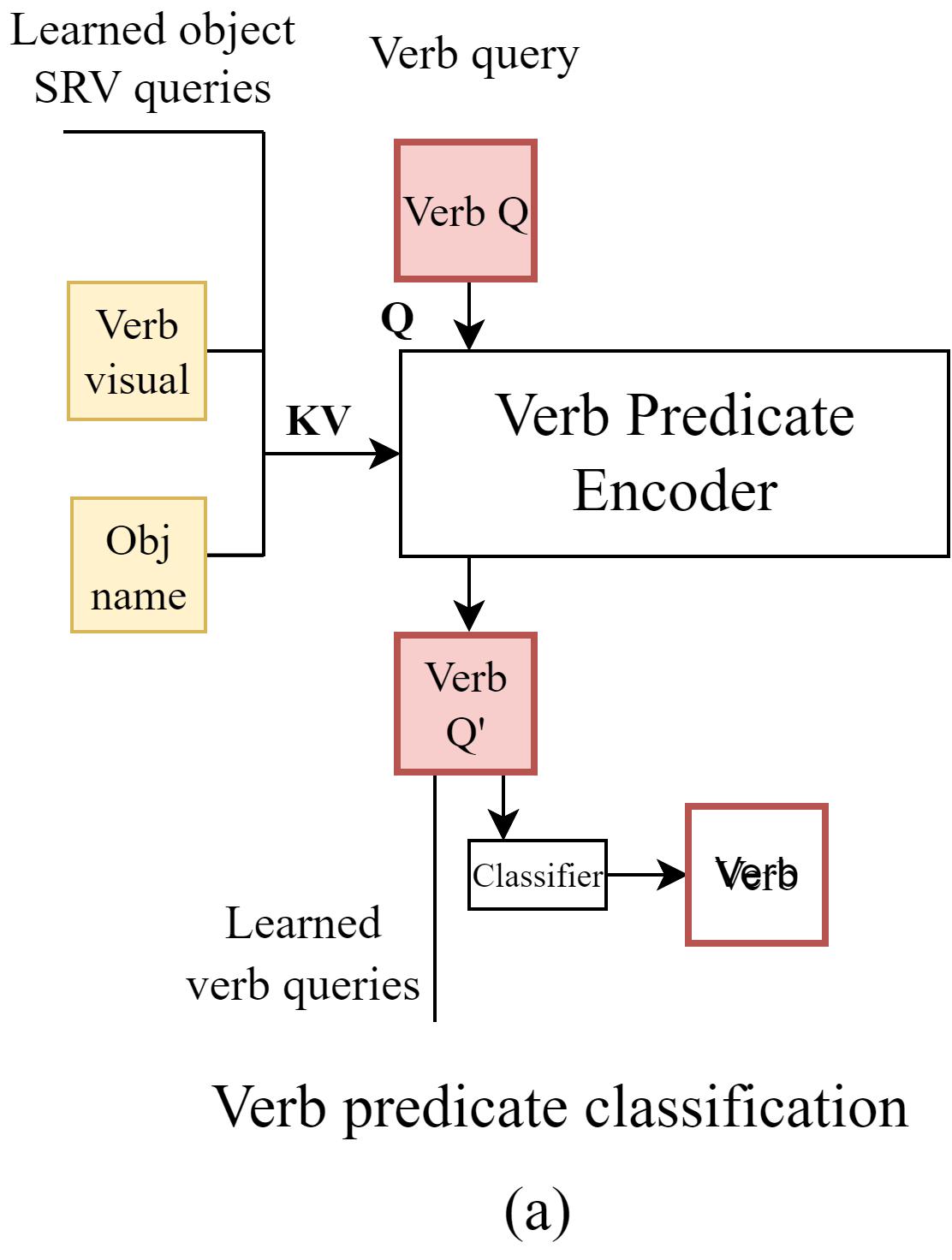}
    \includegraphics[width=0.3\textwidth]{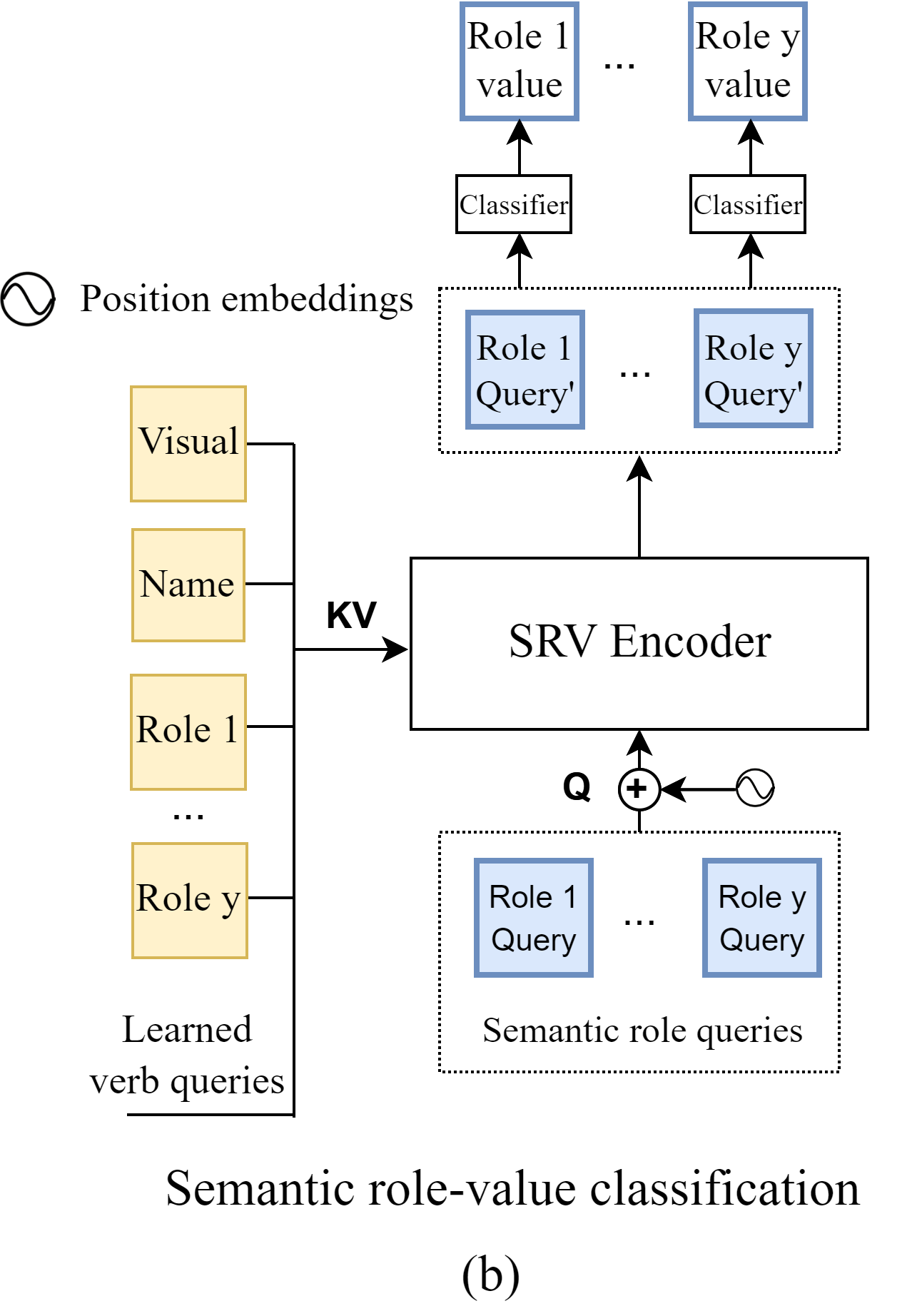}
    \caption{Architectures of verb predicate and SRV encoders.}
    \label{fig:detaild_modules}
\end{figure}

\sloppy
\subsection{Situational scene graph generation}
For a given video frame, the SSG generation task requires the classification of the (1) verb predicate classes and (2) semantic role-values of the person, object instances, 
and verb predicate of the relation instances as follows.

\noindent
\textbf{(1) Verb predicate (VP) classification:}
Given a video frame consisting of a person and class labels of object instances with their bounding boxes as inputs, a model is required to recognise the verb predicate class $\mathcal{R}$ of the relation instances between person and object instances.

\noindent
\textbf{(2) Semantic role-value (SRV) classification:}
Here the model is required to recognize the role-value $v$ of each semantic role of person, object, and verb predicate of relation instances in a given frame as shown in Eq. \ref{eq.value.classification}.
As mentioned earlier, while semantic roles are shared across a category, the values depend on the specific instance of that category.

\subsection{Proposed SSG model: InComNet}
\sloppy
We hypothesize that semantic role-value classification helps predicate classification (in scene graph) and vice versa. 
Based on this, we introduce the InComNet, \textbf{In}teractive and \textbf{Com}plementary \textbf{Net}work, a framework designed for Situational Scene Graph (SSG) generation, which also demonstrates the dataset's utility for this task.
As shown in Fig. \ref{fig:architecture}, InComNet has four stages: (I) object SRV classification, (II) verb predicate classification (III), verb predicate SRV classification, and (IV) person SRV classification. 
The stage (II) classifies the verb predicates in the relation instances between person and object instances while
stages (I), (III) and (IV) classify the semantic role-value of each semantic role of object, verb predicate of the relation instances and person respectively as in Eq. \ref{eq.value.classification}.
In InComNet, we use transformer encoders with cross attention, where learnable query vectors serve as queries, and a sequence of input features act as keys and values.
During training, these learnable queries interact with the features in the input sequence and construct useful learned query representations that can be used for the classifications. 
Accordingly, we employ four such encoders, namely, object SRV encoder, verb predicate encoder, verb predicate SRV encoder and persons SRV encoder corresponding to the InComNet stages (I), (II), (III) and (IV), respectively. The verb predicate encoder's detailed architecture is presented in Fig. \ref{fig:detaild_modules}a. Object, verb predicate and person SRV encoders follow the same architecture as shown in Fig. \ref{fig:detaild_modules}b with the exception that person SRV encoder does not include an intermediate input for learned object queries.

\sloppy
Leveraging the effectiveness of CLIP~\cite{clip} embeddings, we derive visual and semantic representations using CLIP's vision ($\psi_v()$) and text ($\psi_t()$) encoders. 
Additionally, we use a visually prompted frame embedding, employing a new visual prompt named a \textbf{translucent background prompt}. 
In this prompt, we applied a translucent pink overlay to the non-interested regions in the video frame,
allowing the model to pay more attention to the interested region. 
Here the interested region refers to the bounding box region (e.g. in object SRV classification, the prompt is applied to the non-bounding box region of the object while retaining the bounding box region in its original state. Refer to Section \ref{sec:Evaluation of different visual prompts} for more details).
Let us denote the prompted frame by $f^{pr}$ and the corresponding visual feature by $U_{f^{pr}} = \psi_v(f^{pr})$.
Similarly, text embedding of the object class name $U_{O_i}$ where $O_i \in \mathcal{O}$, and text embeddings of the object semantic role names $\{ U_{S_1(O_i)}, U_{S_2(O_i)}, \cdots\}$ are obtained from the CLIP text encoder $\psi_t$.

\noindent
\textbf{Object SRV classification:} 
\sloppy 
In the first iteration, the object SRV encoder receives a sequence of key/values $KV = \{U_{f^{pr}}, U_{O_i}, U_{S_1(O_i)}, \cdots, U_{S_x(O_i)} \}$, where $x$ is the maximum number of semantic roles for an object in the SSG dataset. For objects with fewer than $x$ roles, the remaining embeddings are zero-padded. Queries are formed by $x$ learnable query vectors, producing $x$ role query embeddings $Q_{r}^O = \{q_{r_1}^{O}, \cdots, q_{r_x}^{O}\}$. A linear classifier predicts the correct role-values using $Q_{r}^O$. 
In our semantic frame definition, the positions of roles for objects, verb predicates and person remain consistent across the dataset. Hence, only the output embedding corresponding to the appropriate role is utilized during loss calculation.

\noindent
\textbf{Verb predicate classification:} 
We hypothesize that object semantic roles and values enhance the classification of verb predicates in the relation between a person and object. Thus, the output of the object SRV encoder from stage (I) is used as additional input for classifying verb predicates $R_i$ in stage (II). The input sequence for stage (II) includes the prompted frame $U_{f^{pr}}$, object class embedding $U_{O_i}$, and the learned object role query embeddings $Q_{r}^O$ forming the keys and values $KV = \{U_{f^{pr}},U_{O_i}, Q_{r}^O\}$. A learnable verb predicate query vector is then used to generate the verb predicate query embedding $Q^{R}$.

\noindent
\textbf{Verb predicate SRV classification:} 
The verb predicate SRV encoder receives an input sequence comprising the prompted frame (verb visual features) $U_{f^{pr}}$, the learned verb predicate query embedding $Q^{R}$, the object class name embedding $U_{O_i}$ associated with the verb predicate, and the role name embeddings $\{U_{S_1(R_i)}, \cdots, U_{S_y(R_i)}\}$.  This sequence forms the keys and values $KV = \{U_{f^{pr}}, Q^{R}, U_{O_i}, U_{S_1(R_i)}, \cdots, U_{S_y(R_i)}\}$, while $y$ learnable verb predicate role query vectors generate the corresponding role query embeddings $Q^{R}_r$.
From the second iteration onward, the learned verb predicate query embedding $Q^{R}$ from the predicted verb predicate instances is used as an additional input to stage (I), complementing the object SRV classification, based on the hypothesis that verb predicates assist in classifying object semantic role-values. Consequently, in iteration 2, the input sequence for stage (I) becomes $KV = \{U_{f^{pr}}, U_{O_i}, Q^{R}, U_{S_1(O_i)}, \cdots, U_{S_x(O_i)}\}$.

\sloppy
\noindent
\textbf{Person SRV classification:}
The input sequence for stage (IV) includes the prompted frame  (person visual features) $U_{f^{pr}}$, the text embedding of the person name\footnote{in all cases we use term "person".} $U_P$, and role name embeddings $\{U_{S_1(P)}, U_{S_2(P)}, \cdots, U_{S_z(P)}\}$, forming the keys and values $KV = \{U_{f^{pr}}, U_P, U_{S_1(P)}, \cdots, U_{S_z(P)}\}$ for the person SRV encoder. Queries are generated by $z$ learnable person role query vectors, producing $z$ learned person role query embeddings $Q^{P}_{r}$.
The learned query embeddings $Q^{O}_{r}$, $Q^{R}$, $Q^{R}_{r}$ and $Q^{P}_{r}$ of stages (I), (II), (III) and (IV) are passed through four linear classifiers to generate the final situational scene graph by classifying object semantic role-values, verb predicates, verb predicate semantic role-values and person semantic role-values respectively.

Stages (I), (II), and (III) are iterated $d$ times, with their losses calculated by summing cross-entropy losses across all iterations. The loss for stage (IV) is computed using cross-entropy loss between predicted and ground truth role-values.

\subsection{SSG dataset}
\sloppy
We introduce \textbf{S}ituational \textbf{S}cene Graph (SSG) dataset to facilitate the proposed human-action representation of situational scene graphs. 
SSG dataset is built on Action Genome's ~\cite{ag} spatio-temporal scene graphs and Action Genome is built on Charades~\cite{charades}.
However, unlike the Action Genome whose goal is to decompose Charades' video-level actions
by annotating person, objects and their pairwise relationships, SSG's goal is to provide precise structures to further elaborate these action partonomies using semantic roles and role-values. 
Moreover, unlike VidSitu \cite{sadhu2021visual} which annotates a single action with semantic roles
for every 2-second clip, SSG dataset provides frame-level situations in a more granular way than VidSitu.
An analysis of it in comparison to other publicly available video understanding datasets is given in Table \ref{tab:datasets}.
Among the main components of situational scene graphs, persons, object instances and verb predicates of the relation instances are already annotated in the Action Genome. We provide additional annotations for semantic roles and values for the above semantic entities. 
 
\sloppy
\noindent
\textbf{Composing the dataset:} 
In the Action Genome, each person-object instance has three types of relationships: attention, spatial and contacting. We provide semantic role-values for verb relationships which we refer as verb predicates because these verb predicates describe various ways that a person interacts with an object to execute a higher-level action as shown in \ref{fig:method}. 
Out of its 17 verb predicate classes, we decided to remove \emph{other relationship} as it is too vague to represent an action as well as is particularly challenging to decompose into specific semantic roles and values.
The remaining 16 verb predicates, however, effectively represent situations involving ongoing actions when combined with semantic roles and values.
We did not perform any filtration to the Action Genome objects and thus it contains 35 object classes and a person class.
Next, following the concept of semantic roles in FrameNet~\cite{fillmore2003background} and previous work ~\cite{yatskar2016situation}, we assigned semantic roles for the person, object and verb predicate classes.  
The raw semantic roles, particularly those related to objects and person, were further refined to ensure their applicability in visually describing a scene.
We developed a web-based application software called Semantic Role Labelling Tool to facilitate the annotation process. Annotations were carried out over a period of three months
by students with a Computer Science background.
Refer Section \ref{sec:ssg_dataset} for more details about the annotation process, the labelling tool, distribution of the dataset and further dataset statistics.

\section{Experiments}
\label{sec:Experiments}


\noindent
\textbf{Experimental settings:} 
\sloppy
We use our SSG dataset for the SSG generation task.
We use the translucent background prompt in all the visual prompt-related experiments and
trained the InComNet with a learning rate of 0.001, and ExponentialLR scheduler with Adamax optimizer, on a 49 GB NVIDIA RTX 6000 GPU.

\noindent
\textbf{Evaluation metrics:} 
For the verb predicate classification task,  
we use the metric \textbf{accuracy} (Acc) \cite{yatskar2016situation} under "with constraint" predicate classification (PREDcls)~\cite{ag} setting.
Further, since semantic role-value classification is similar to the semantic role labelling in conventional verb-based situation frames, we adapt existing metrics \textbf{value} and \textbf{value-all} \cite{pratt2020grounded} along with a new metric called \textbf{value-two}.
However, unlike in \cite{yatskar2016situation, pratt2020grounded}, we have only single annotation per record. Therefore, we assess the value by determining whether the model accurately predicts at least one role-value for a given role out of all the roles associated with that particular semantic entity. The value-all metric evaluates if the model can accurately predict all the semantic role-values, while the value-two metric checks if at least two semantic role-values are correctly predicted out of all the roles.
We use top-1 setting for verb predicate SRV classification where semantic role-values are considered incorrect if the predicted verb is incorrect.  
Since we do not perform object detection, we use the ground truth setting for object and person SRV classifications
assuming ground truth object/person to be known \cite{yatskar2016situation, pratt2020grounded}.
For SRV classification, we also introduce a new metric called \textbf{role-based accuracy} (role-based acc.) to understand the role-based performance. It is calculated by first determining the accuracy for each role separately, and then averaging across all roles.

\subsection{Situational Scene Graph Generation}

\paragraph{Baselines:} We compare the performance of InComNet over six baselines under three categories: (1) CLIP zero-shot inference \cite{clip}, (2) CLIP linear probing \cite{clip} and (3) VILA inference \cite{lin2024vila}.
In CLIP linear probing \cite{clip}, we attach a linear classifier on top of the frozen CLIP vision encoder for each object SRV, verb predicate, verb predicate SRV and person SRV classifications.
In VILA inference, we prompt VILA-1.5-3B model in interleaved image-text VQA MCQ style with MCQ questions generated using the SSG annotations. Under this category we showcase both zero-shot and impact of fully fine-tuning an LVLM like VILA on the SSG task. 
The option space for verb predicate and SRV MCQ questions includes the set of verb predicate classes and unique set of role-values associated with each semantic role in the SSG dataset respectively.
More details about fine-tuning settings for VILA, including the prompts used are given in the Section \ref{sec:vila_sft}.

\paragraph{InComNet models:} 
We have three InComNet variants as (1) InComNet-224 (base model), (2) InComNet-336-Frozen and (3) InComNet-336-FT. We use ViT-B-32 and ViT-L-14-336 CLIP vision and text encoders to extract image and text feature embeddings in (1) and (2) respectively, while in (3), we fine-tune ViT-L-14-336 CLIP model on SSG dataset and then use this model to extract the required embeddings. We perform all ablation studies and hyper-parameter analysis etc. using the InComNet-224 base model.
When fine-tuning the Clip ViT-L-14-336 model on our SSG dataset, we created a detailed description about each video frame with the SSG annotations. Refer \ref{sec:fine-tuning clip} for more details about CLIP fine-tuning on SSG dataset.
We refer InComNet's individual tasks as the scenario where its sub-tasks- object SRV, verb predicate and verb predicate SRV are trained independently without inter-task communication.

Table \ref{tab:main_results} shows the performance comparison of InComNet with baselines over the SSG dataset.
Our analysis reveals following key insights:
\textbf{(1) Superior performance of InComNet-336-FT}: InComNet-336-FT outperforms VILA1.5-3b-FT by an average margin of 15.5\% and CLIP linear probing-336-FT model by an average margin of 34.3\%;
\textbf{(2) Enhances performance through joint and iterative training}: Overall performance of object SRV, verb predicate and verb predicate SRV classifications improves by 4\% when they are iteratively and jointly trained with complementary exchange of input-output information from other sub-tasks than training each sub-task individually. This demonstrates the synergistic benefits of our approach, which validates our hypothesis that semantic role-value classification and predicate classification mutually enhance each other;
\textbf{(3) Impact of higher image resolution}: Increasing the image resolution from 224 to 336 pixels in the InComNet-336-Frozen model significantly enhances its capability to capture fine-grained details, resulting in an average performance improvement of 5.6\% over the InComNet-224. This underscores the importance of higher resolution for improved model performance in detailed image analysis;
\textbf{(4) Benefits of finetuning CLIP features:} Finetuning CLIP vision and text encoders for specific tasks like ours would lead to notable performance gains. Specifically, the ViT-L-14-336-FT model shows a 7.2\% improvement over the ViT-L-14-336-Frozen model;
\textbf{(5) Challenges and opportunities with SSG:} SSG task presents a notable challenge to foundational models like VILA when prompted with large option space. While the value and value-two metrics suggest that the model is capable of identifying one or two dominant properties, the lower performance in the value-all metric highlights that it struggles to accurately identify the majority of the properties associated with people, objects, and verb relationships. 
This underscores the potential for continued research, positioning SSG as a promising area for exploration by LVLMs. 

\begin{table*}

  \centering
  \begin{center}
  \resizebox{1\textwidth}{!}{
  \begin{tabular}{|l|c|c|c|c|c|c|c|c|c|c|c|c|c|}
    \hline
    \multirow{2}{*}{Method} 
    & \multicolumn{4}{c|}{Object SRV classification} 
    & \multirow{2}{*}{\shortstack{Verb predicate \\ classification acc.}}
    & \multicolumn{4}{c|}{Verb predicate SRV classification} 
    & \multicolumn{4}{c|}{Person SRV classification}\\

     \cline{2-5}  \cline{7-14} 

    & \multicolumn{1}{c|}{Role-based acc.} & \multicolumn{1}{c|}{Value} &  \multicolumn{1}{c|}{Value-two} & \multicolumn{1}{c|}{Value-all} & & \multicolumn{1}{c|}{Role-based acc.} & \multicolumn{1}{c|}{Value} & \multicolumn{1}{c|}{Value-two} & \multicolumn{1}{c|}{Value-all} & \multicolumn{1}{c|}{Role-based acc.} & \multicolumn{1}{c|}{Value} & \multicolumn{1}{c|}{Value-two} & \multicolumn{1}{c|}{Value-all}\\

\hline						
CLIP-ViT-B-32  zero-shot inference \cite{clip}	& 1.5    & 6.9	& 0.2	& 0.0	& 8.9	& 0.4	& 0.1	& 0.0	& 0.0	& 0.1	& 0.2	& 0.0	& 0.0 \\

\hline

CLIP linear probing-224 \cite{clip} &  11.0	& 52.7	& 2.9	& 1.1	& 40.4	& 3.3	& 39.0	& 0.0	& 0.0	& 17.3& 83.2 &	3.3& 0.0 \\
CLIP linear probing-336-Frozen \cite{clip}  &  11.2	& 54.4	& 3.5	& 1.1	& 40.1 & 3.4 &	38.5 &	0.0	& 0.0 &	19.1 &	89.7 &	3.6 &	0.0 \\
CLIP linear probing-336-FT \cite{clip}  & 11.2	& 54.5	& 3.6	& 1.8 &	47.5 & 3.8	& 46.1	& 0.0	& 0.0	& 19.2	& 90.3	& 4.9	& 0.0 \\

\hline
VILA1.5-3b-z-shot (MCQ all options) \cite{lin2024vila}	& 11.2	& 40.4	& 8.8	& 0.0	& 19.6	& 14.1	& 19.1	& 13.0	& 1.7	& 35.0	& 88.2	& 57.2	& 0.0\\
VILA1.5-3b-FT (Instruction-tuned MCQ all options) \cite{lin2024vila}	& 21.0	& 69.2	& 26.4	& 0.2	& 61.3	& 35.3	& 59.6	& 35.4	& 3.2	& \underline{44.4}	& 95.7	& 74.6	& 1.0\\

\hline	
InComNet's individual tasks (no iterations/feedback) & 45.5	& 86.2	& 54.0	& 5.7 & 66.9 & 26.9	& 65.3	& 63.4	& 28.3 & - & - & - &  \\
InComNet-224	& 46.8	& 86.6	& 55.0	& 5.9	& 67.8	& 29.4	& 66.2	& 64.7	& 32.0	& 39.8	& 97.4	& 77.4	& 1.6 \\
InComNet-336-Frozen	& \underline{47.4}	& \underline{87.6}	& \underline{58.3}	& \underline{6.8}	& \underline{68.5}	& \underline{36.3}	& \underline{66.7}	& \underline{64.8}	& \underline{32.6}	& 42.2	& \underline{98.0}	& \underline{77.7}	& \underline{1.8}\\
InComNet-336-FT	& \textbf{49.6}	& \textbf{89.5}	& \textbf{62.1}	& \textbf{7.8}	& \textbf{70.8}	& \textbf{38.4}	& \textbf{69.0}	& \textbf{67.4}	& \textbf{38.7}	& \textbf{49.3}	& \textbf{98.5}	& \textbf{85.8}	& \textbf{1.9} \\

  \hline
  \end{tabular}%
  }
  \end{center}
  \caption{Performance of InComNet for SSG generation task on SSG dataset. The bold and underlined font show the best and the second best result respectively.}
  \label{tab:main_results}
\end{table*}%

CLIP inference results are sub-optimal, likely due to its training focus on tasks like image classification, which may introduce a bias towards salient objects within the image. Furthermore, the CLIP model is not tailored for structured prediction similar to our InComNet.
In contrast, SSG requires classifying \emph{multiple objects, verb predicates, the person and their semantic role-values in a single video frame}, making the accurate mapping of these elements more challenging.
Further ablation studies about InComNet, its iterative refinement of the results including visualizations, per-class and per-role performances of verb predicate and SRV classification can be found in Section \ref{sec:incomnet_settings}. 
Implementation details and hyper-parameter analysis of the InComNet are given in Section \ref{sec:hyperparamters}.
Translucent background prompt proved to be the most effective visual prompt for the SSG task compared to existing visual prompts \cite{zhang2023fine, bahng2022exploring} and more details about prompt evaluation can be found in Section \ref{sec:Evaluation of different visual prompts}.

\subsection{Applications of situational scene graphs}

We now demonstrate how our unified SSG representation can benefit both situation recognition and predicate classification tasks in complementary ways. These tasks are paired with the key building blocks of our representation i.e. situation frames \cite{yatskar2016situation} and ST scene graphs \cite{ag} respectively.

\sloppy
\noindent\textbf{(1) Situation recognition on SSG dataset:}
Since situation frame is a key foundational component of our SSG representation, situation recognition task could also be tested on the SSG dataset.
The only difference between typical situation recognition problem on datasets such as imSitu \cite{yatskar2016situation} and situation recognition on SSG dataset is that, in imSitu there is only one activity verb and its associated semantic roles and values whereas, the SSG dataset includes multiple activity verbs, each paired with semantic roles and values corresponding to multiple concurrent actions within a single video frame. Therefore, in SSG situation recognition problem, the models should be able to classify all the verb predicates and their associated semantic role-values in a video frame which adds another layer of complexity. Accordingly, we purpose the SOTA situation recognition models i.e. Clipsitu also to predict all the verb predicates and their semantic role-values.

Thanks to our SSG representation, as shown in Table \ref{tab:sr_performance}, our InComNet models consistently outperform clipsitu models and VILA models, achieving average performance improvements of 31.1\% and 18.2\%, respectively, across all metrics. 
Furthermore, it should be noted that, our unified representation enables the simultaneous prediction of verbs and their semantic role-values within a single pipeline, fostering a complementary exchange of information and inter-task communication on verb predicate and SRV classification. In contrast, both the Clipsitu and VILA models depend on a two-stage prediction approach, where the two tasks are handled separately.

\begin{table*}
  \begin{center}
  \resizebox{1\textwidth}{!}{
  \begin{tabular}{|l|c|c|c|c|c|c|c|c|}
    \hline
    
    \multirow{2}{*}{Method} 
    & \multirow{2}{*}{\shortstack{Top-1 verb \\ accuracy}}
    & \multicolumn{3}{c|}{Top-1 verb predicate SRV classification} 
    & \multicolumn{3}{c|}{GT verb predicate SRV classification} 
    \\
    \cline{3-8}

    & &  \multicolumn{1}{c|}{Value} & \multicolumn{1}{c|}{Value-two} & \multicolumn{1}{c|}{Value-all} & \multicolumn{1}{c|}{Value} & \multicolumn{1}{c|}{Value-two} & \multicolumn{1}{c|}{Value-all} \\
    
    \hline

    Clipsitu MLP \cite{clipsitu}		& \multirow{3}{*}{41.5}	& 40.1	& 30.7	& 0.8	& 97.1	& 55.3	& 9.0\\
    Clipsitu TF	 \cite{clipsitu}       & 	& 40.0	& 33.0	& 2.5	& 97.1	& 57.0	& 8.5\\
    Clipsitu XTF \cite{clipsitu}		& 	& 39.9	& 33.0	& 2.2	& 97.1	& 54.9	& 6.2 \\

    \hline
    
    VILA1.5-3b-z-shot \cite{lin2024vila}	& 19.6	& 19.1	& 13.0	& 1.7	& 96.8	& 54.3	& 5.5\\
    VILA1.5-3b-FT \cite{lin2024vila}	    & 61.3	& 59.6	& 35.4	& 3.2	& \textbf{98.2}	& 82.0	& 30.4\\
    
    \hline
    
    InComNet-224    	& 67.8	& 66.2	& 64.7	& 34.0	& 97.2	& \underline{94.0}	& \underline{51.9}\\
    InComNet-336-Frozen	& \underline{68.5}	& \underline{66.7}	& \underline{64.8}	& \underline{32.6}	& \underline{97.3}	& 92.7	& 38.5	\\
    InComNet-336-FT		& \textbf{69.8}	& \textbf{68.2}	& \textbf{66.9}	& \textbf{41.2}	& \textbf{98.2}	& \textbf{94.8}	& \textbf{58.1}	 \\

  \hline
  \end{tabular}%
  }
  \end{center}
  \caption{Performance of SOTA task-specific situation recognition models and foundation models on SSG dataset for situation recognition. The bold and underlined font show the best and the second best result respectively. 
  }
  \label{tab:sr_performance}
\end{table*}%

\sloppy
\noindent\textbf{(2) Predicate classification with situational scene graphs on Action Genome dataset:}
In our SSG dataset, we provided annotations for only 8\% of the Action Genome dataset. In this experiment, we trained the InComNet model for predicate classification (under PredCLS setting and with-constraint strategy) on this limited 8\% and used transfer learning to infer object and verb predicate semantic role-value annotations for the entire Action Genome test set during inference. This approach enabled effective evaluation of the predicate classification problem across the entire Action Genome dataset with the help of its 8\% SSG annotations.
It is worth to mention that, while others have used the entire Action Genome dataset during training, our model has used only 8\% Action Genome frames for which we have provided the SSG annotations. 
We attach three linear classifiers for the InComNet's verb predicate encoder for the three relationship types in Action Genome.
The results in Table \ref{tab:sg_performance} demonstrate that we can achieve comparable performance with SOTA methods with just 8\% data.

\begin{table}[t!]
  \begin{center}
  \resizebox{0.4\textwidth}{!}{%
  \begin{tabular}{|l|c|c|c|}

    \hline
    Method &R@10 & R@20 & R@50  \\
    \hline

    VRD \cite{lu2016visual} & 51.7 & 54.7 & 54.7  \\
    M-FREQ \cite{zellers2018neural} &    62.4 & 65.1 & 65.1 \\
    MSDN \cite{li2017scene} &  65.5 & 68.5 & 68.5 \\
    VCTree \cite{tang2019learning} &  66.0 & 69.3 & 69.3  \\
    RelDN \cite{zhang2019graphical} &  66.3 & 69.5 & 69.5  \\
    GBS-Net \cite{lin2020gps} &  66.8 & 69.9 & 69.9   \\

    STTran TPI \cite{wang2022dynamic}  & \underline{69.7} & 72.6 & 72.6  \\
    APT \cite{li2022dynamic} &  69.4 & \textbf{73.8} & \textbf{73.8}  \\
    TD$^{2}$-Net(P) \cite{lin2024td} &  70.1 & - & 73.1 \\
    STTran \cite{sttran} &  68.6 & 71.8 & 71.8\\
    DSG-DETR \cite{feng2023exploiting} & 68.4 & 71.7 & 71.7\\
    TR$^{2}$ \cite{wang2023cross} & \textbf{70.9} &  \textbf{73.8} &  \textbf{73.8} \\
    TEMPURA \cite{nag2023unbiased} &  68.8 & 71.5 & 71.5 \\
    CLIP zero-shot \cite{clip} & 19.4 &	20.9 &	20.9 \\

    \hline

    InComNet-224 & 69.1 & 69.2 & 69.2 \\
    InComNet-336-Frozen & 67.8 & 70.8 & 70.9 \\
    InComNet-336-FT & 69.4 & \underline{72.7} & \underline{72.7} \\

    \hline
  \end{tabular}%
  }
\end{center}
    \caption{Performance of SOTA predicate classification models on Action Genome dataset. \textbf{Our method uses only 8\% AG frames for training}. The bold and
underlined font show the best and the second best result respectively.}
    \label{tab:sg_performance}
\end{table}%

\subsection{Reasoning on human-centric situations}

Next, we evaluate the usability of our SSG representation for human-centric situation understanding by leveraging the zero-shot image-text interleaved MCQ inference capabilities of the VILA 1.5-3B model \cite{lin2024vila}.
For this experiment, we generate 139K+ MCQ questions using the annotations in the SSG test set each having \textbf{four} options. Thus, the test questions set comprises of 12K verb predicate questions, 38K verb predicate SRV questions, 64K object SRV questions and 25K person SRV questions. Performance is measured using accuracy.
The results are shown in Table \ref{VQA}. In (3) and (4) the predicted/ GT SSG graph in text format is incorporated as an in-context reference. 
The results demonstrate that, incorporating the predicted SSG graph as an in-context yields an average of 10\% improvement in accuracy while GT SSG graph leads to 24\% marking a certain upper bound on performance when using graph data.
This underscores the value of explicitly representing the elements in situations using the SSG graph structure for improved human-centric situated understanding and reasoning.
Additionally, it highlights the potential avenues for enhancing future SSG models on this area.

\begin{table}[t!]
\centering
\resizebox{0.9\textwidth}{!}{%
\begin{tabular}{|l|c|c|c|c|}
\hline

{\shortstack{Input  }}  & {\shortstack{Obj SRV  \\ questions}}  & {\shortstack{Verb predicate \\ questions}}  & {\shortstack{Verb predicate \\ SRV questions}}  & {\shortstack{Person SRV \\ questions}}  \\ \hline
(1) Question & 51.1 & 54.1 & 60.3 & 25.1 \\ 
(2) Image + question & 53.9 & 60.1 & 70.6 & 46.2 \\ 
(3) Image + predicted graph + question & \underline{60.0} & \underline{70.8} & \underline{75.1} & \underline{50.0} \\ 
(4) Image + GT graph + question & \textbf{76.9} & \textbf{85.5} & \textbf{78.5} & \textbf{71.8} \\

\hline
\end{tabular}%
  }
\caption{SSG VQA on VILA \cite{lin2024vila}. The bold and
underlined font show the best and the second best result respectively.}
\label{VQA}
\end{table}

\section{Challenges and outlook}

\paragraph{Challenges}

The SSG tasks are particularly challenging due to the need for (1) strong visual cues to identify the attributes of smaller objects (e.g., the material of a dish), subtle variations in human poses (e.g., sitting vs. leaning), precise object locations (e.g., a laptop on a table or on someone's lap), recognizing specific body parts involved in actions, and the tools used to perform those actions. 
Additionally, (2) situational common sense is essential for understanding semantic role-values for certain roles, such as object affordances (e.g., a doorknob's affordance when opening a door is unlocking, while it is locking when closing). Finally, (3) unlike problems such as typical situation recognition which requires recognising one salient activity verb and its semantic role-values, the SSG task particularly requires recognising multiple verb predicates, multiple objects, the person and their semantic role-values from a single video frame that correspond to multiple concurrent actions. 

\paragraph{Outlook}
SSG dataset has shown potential to enhance performance in existing vision tasks such as situation recognition, predicate classification and human-centric situation reasoning by leveraging its detailed annotations.
Beyond these tasks, we also foresee the potential of SSG in pushing the current VLMs towards a more granular level of video perception and reasoning specially in the following areas:
\textbf{(1) Video QA}: 
While current video questions answering datasets such as STAR \cite{star} and AGQA \cite{grunde2021agqa} lack fine-grain details about entities involved in the situations, the inclusion of semantic roles and values of those entities could be crucial for enhancing the situated reasoning capabilities of the VLMs, which we list as a future work;
\textbf{(2) Dense video captioning}: SSG annotations can also be utilized for dense video captioning and serve as a benchmark for video-description pre-training and evaluation. We also list this as a potential future direction;
\textbf{(3) Video generation}: recent work explores image/video generation from structured representations \cite{cong2023ssgvs}. Given the fine-grained semantic properties embedded in situational scene graphs, we anticipate that future SSG models could enable the generation of videos from situational scene graphs, opening new avenues in video synthesis.
%

\section{Conclusion}
\label{sec:Conclusion}

We introduce situational scene graphs, a unified graph representation which can represent both human-object relations and semantic properties of the entities involved in a human centric situation.
In doing so, we also introduce a new task called situational scene graph generation accompanied by a new dataset, SSG and a new model to address this task. 
Finally we demonstrate the utility of our SSG representation for situation recognition, predicate classification and human-centric situation understanding.
The need for extensive annotations to construct SSG representation may be seen as a limitation. Yet, the advancement of structured semi-supervised learning approaches may allow us to generate better annotations unleashing the full potential of SSGs for human-centric situation understanding and reasoning.
\newline
\textbf{Acknowledgment} This research/project is supported by the National Research Foundation, Singapore, under its NRF Fellowship (Award\# NRF-NRFF14-2022-0001). This research is also supported by funding allocation to B.F. by the Agency for Science, Technology and Research (A*STAR) under its SERC Central Research Fund (CRF), as well as its Centre for Frontier AI Research (CFAR).

\section{Supplementary material}
\label{sec:suppl}

This supplementary material provides further details about the
SSG dataset
in Section \ref{sec:ssg_dataset}, 
fine-tuning settings of VILA \cite{lin2024vila} for SSG dataset in Section \ref{sec:vila_sft},
fine-tuning settings of CLIP \cite{clip} for SSG dataset in Section \ref{sec:fine-tuning clip},
detailed settings and further experiments on InComNet including performance of InComNet across iterations, visualizations for InComNet's situational scene graph predictions and per-class and per-role performance of InComNet in Section \ref{sec:incomnet_settings},
hyper-parameters of InComNet in Section \ref{sec:hyperparamters},
evaluation of different visual prompts for situational scene graph generation task in Section \ref{sec:Evaluation of different visual prompts},
predicate classification related further experiments in Section \ref{sec:pred_cls},
and 
action recognition with situational scene graphs in Section \ref{sec:action_recognition}
which could not be included in the main paper due to the limited space.


\paragraph{Human-centric vs non-human-centric situations}
\hfill\break
In situation recognition literature, a situation can involve either a human or an animal as agent performing the action such as bear \cite{yatskar2016situation}.
We define \textbf{Human-centric situations} as those where a human is the central figure encompassing multiple concurrent actions performed by the human, human-object relationships, and the semantic properties of the involved entities.   
Therefore, in our situational scene graphs, we always focus on scenarios involving a person, thus making these situations human-centric by definition.
\hfill\break
\hfill\break
We will make the code and our SSG dataset publicly available upon the acceptance of this paper.


\subsection{SSG dataset} 
\label{sec:ssg_dataset}

\paragraph{Existing video representations and required annotations:}
Table \ref{tab:representations} outlines the existing video representations, required annotations and related benchmarks.
In our SSG representation, we leverage the strengths of both spatio-temporal scene graphs and situation frames to introduce a unified graph representation which can represent both human-object relations and the semantic properties of the entities involved in a
human-centric situation with multiple concurrent actions.

\begin{table*}[t]

  \begin{center}
  
  \resizebox{1\textwidth}{!}{
  
  \begin{tabular}{|@{}l|l|l@{}|}
    \hline
    Representation & Annotations & Dataset\\
    \hline
    Spatio-temporal scene graph (2020) & Actions, persons, objects, relationships, bounding boxes (bboxes) & Action Genome \cite{ag}\\ 

    3D semantic scene graph (2020) & Relationships, attributes, object class hierarchies & 3DSSG \cite{wald2020learning} \\
    
    Video semantic role labeling (2021) & Verbs, \textbf{verb SR frames}, \textbf{verb SRV pairs}, co-referencing, event relations, temporal segments & VidSitu \cite{sadhu2021visual} \\
    
    Situation abstraction and reasoning (2021) & Actions, objects, relationships, questions, options, answers, bboxes & STAR \cite{star} \\
    
    Panoptic video scene graph (2023) & Objects and relationships with panoptic segmentation & PVSG \cite{yang2023panoptic} \\
    
    \hline

    \multirow{2}{6cm}{Situational scene graph  (ours)}  & Actions, persons, objects, verb predicates, \textbf{person SR frames}, \textbf{object SR frames} & \multirow{2}{4cm}{SSG (ours)}\\
    
    &  \textbf{verb predicate SR frames}, \textbf{person SRV pairs}, \textbf{objects SRV pairs}, \textbf{verb predicate SRV pairs}, bboxes &  \\

  \hline
  \end{tabular}%
  }
  \end{center}
  \caption{A non-exhaustive summary of video representations, required annotations and benchmarks. 
  SRV pairs refers to semantic role-value pairs.
  }
  \label{tab:representations}
  
\end{table*}

\paragraph{SSG dataset annotation process:} 
We developed a web-based application software called Semantic Role Labeling Tool to facilitate the annotation process. 
Annotators were asked to discard the low-quality videos, videos that include more than one person, and frames with incorrectly bounded objects. Videos with more than one person were discarded because in such videos the bounding box of the person who performs the action might shift to other visible person instances in the video. It is noted that, Action Genome has Faster R-CNN detected person bounding boxes.
Further, objects which have been incorrectly bounded (probably due to annotation errors) were also eliminated. 
Since the semantic roles are predefined, annotating a frame is a value-filling task by choosing role-values from a predefined list of possible values. 
Annotators were asked to choose the most appropriate value for a role and if such a value was unavailable in the list, they could add their own values with which the possible value lists were progressively updated.
All the role-values were kept mandatory and if a user is unsure about a particular role-value, the user can select the "unsure" field to avoid entering inaccurate data.

\paragraph{Semantic role labelling tool:} 
The tool was developed using React, Express, and TypeORM frameworks with MySQL. Even though many open source and commercial-off-the-shelf software~\cite{v7,roboflow,cvat,sc} are available for dataset annotation tasks, to the best of our knowledge, there does not exist any software that caters to dataset annotation for situation recognition and scene graph generation. This software not only can be used to provide situation annotation for videos but also can be easily adapted for image data as well. 
We will make the code and the user manual on how to use and adapt the tool publicly available once the paper is published. This tool also created a better platform for us to scale up the dataset at any time.

\begin{figure}[t]
  \begin{minipage}{0.45\textwidth}
    \begin{center}
    \includegraphics[width=\textwidth]{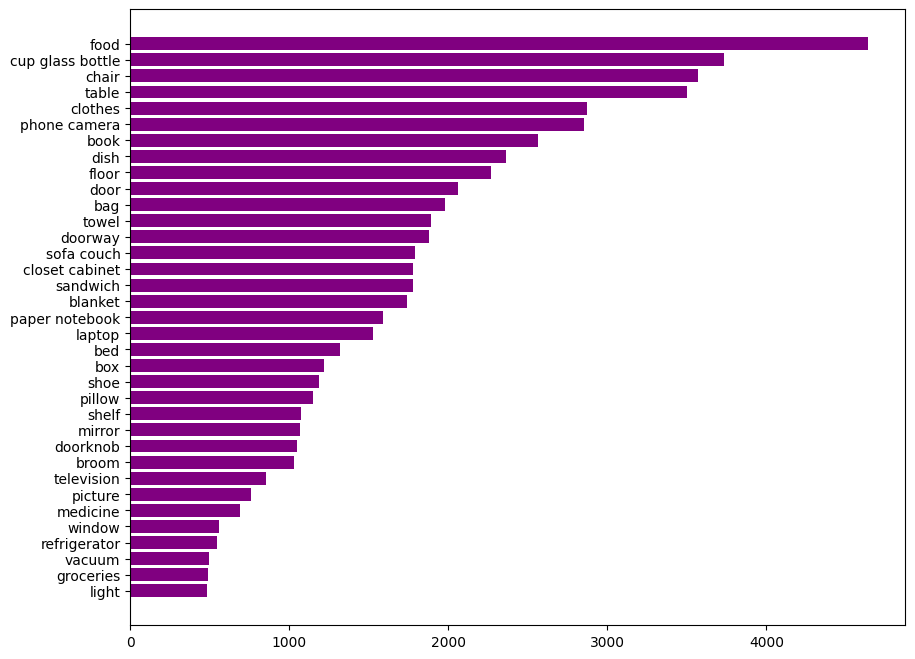}
    \end{center}
  \caption{Distribution of object occurrences in SSG dataset.}
  \label{fig:obj_distribution}
  \end{minipage}\hfill
  \begin{minipage}{0.45\textwidth}
    \begin{center}
    \includegraphics[width=\textwidth]{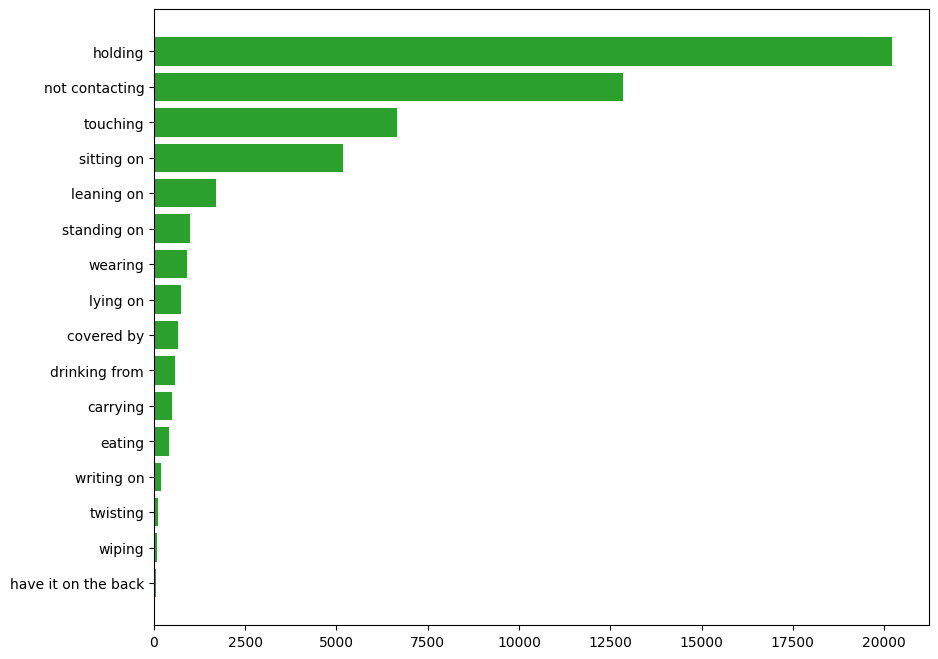}
    \end{center}
    \caption{Distribution of verb predicate occurrences in SSG dataset.}
    \label{fig:rel_distribution}
  \end{minipage}
\end{figure}

\begin{figure}[t]
  \begin{minipage}{0.45\textwidth}
    \begin{center}
    \includegraphics[width=\textwidth]{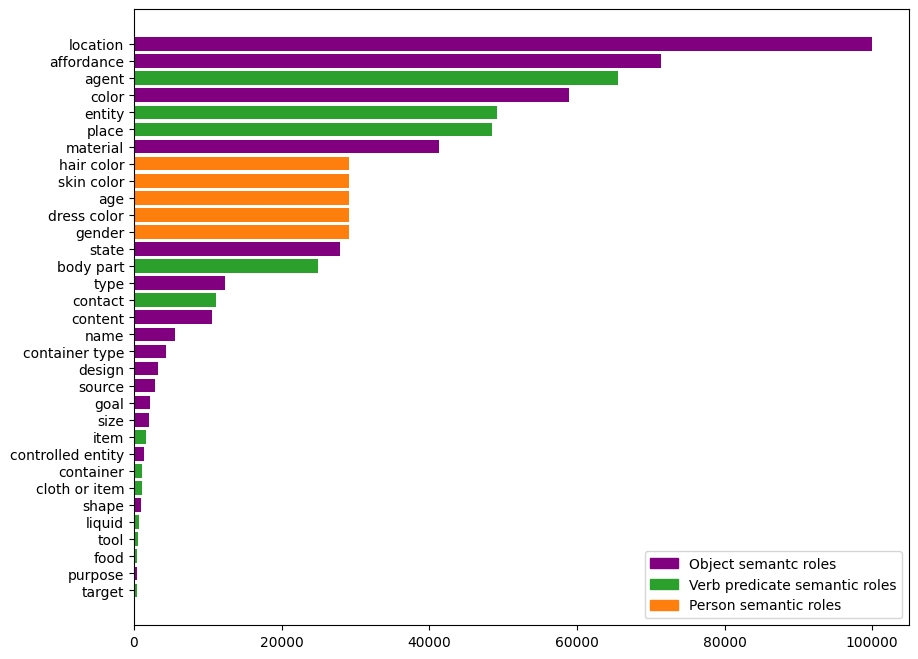}
    \end{center}
    \caption{Distribution of semantic role-value occurrences for person, object and verb predicate semantic roles.}
    \label{fig:role-value_distribution}
  \end{minipage}\hfill
  \begin{minipage}{0.45\textwidth}
    \begin{center}
    \includegraphics[width=\textwidth]{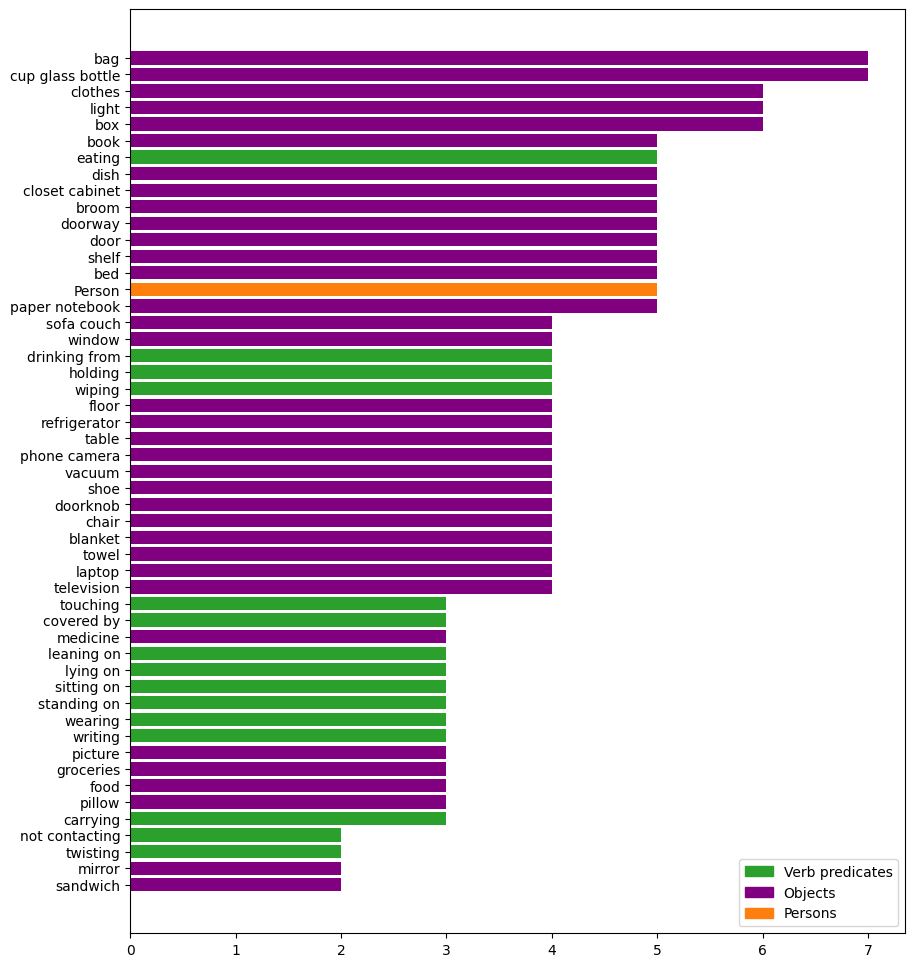}
    \end{center}
  \caption{Distribution of semantic role occurrences for person, objects and verb predicates.}
  \label{fig:entity-role_distribution}
  \end{minipage}
\end{figure}

\paragraph{Distribution of the SSG dataset:}
As shown in Fig. \ref{fig:obj_distribution} and Fig. \ref{fig:rel_distribution}, SSG dataset retains the long-tailed object and verb predicate distributions from Action Genome dataset. Similarly, Fig. \ref{fig:role-value_distribution} illustrates that the semantic role-value (SRV) occurrences of the semantic entities also follow a long-tailed distribution. However, even with such a distribution almost all the semantic entities have at least 0.5K instances. Further, as shown in Fig. \ref{fig:entity-role_distribution}, each semantic entity is associated with a minimum of two and a maximum of seven semantic roles.

\paragraph{Additional dataset statistics:}
The SSG dataset has 25,541 video frames, 
7737.62 semantic roles per object (on average), 
10,266.68 semantic roles per verb predicate, 2.36 objects per video frame, 2.03 verb predicates per video frame, 1.16 verb predicates per object,
10.6 object SRV pairs per video frame,
6.43 verb predicate SRV pairs per video frame,
5 person SRV pairs per video frame,
and 3.41 actions per video frame.


\subsection{Fine-tuning VILA on the SSG dataset for situational scene graph generation task}
\label{sec:vila_sft}

There are several ways of adapting Large Vision Language Models (LVLMs) for other vision-language tasks such as adding linear classifiers on top of image encoders \cite{sung2022vl, yang2023aim, lin2022frozen}, in-context learning \cite{verma2024effectively} and prompting in VQA style \cite{lin2024vila}. 
\textbf{We choose interleaved image-text VQA MCQ style as we plan to extend our SSG dataset towards a video-QA dataset as one of our immediate future works.} 
Accordingly, to comply the SSG tasks with VILA prompting, we created a VQA dataset with 475k training and 140K testing questions using the SSG annotations which specifically focus on verb predicates and SRV of person, objects and verb predicates of human-object relations. 
The question formats we used for the SSG tasks are given in Table \ref{tab:vila_prompts} and the statistics about the questions that we generated are given in Table \ref{tab:SSG-VQA-subset}. These questions were automatically generated from the SSG annotations using the question formats.
Accordingly, the VILA1.5-3b-z-shot model was evaluated using the VQA test set. VILA1.5-3b-FT model with CLIP ViT-L-14-336 vision encoder was instruction-tuned using the VQA train set on a single 49GB NVIDIA RTX A6000 node with four GPUs with batch size 16 and learning rate 1e-4.

For the experiment in Table 2 of the main paper, we defined the option space for questions to align with the SSG tasks. Specifically, the option space for questions related to verb predicates consisted of the verb predicate classes in the SSG dataset. For questions concerning SRV, the option space was defined as the unique set of role-values corresponding to each semantic role in the SSG dataset. Hence, the number of options ranged from a minimum of two to a maximum of 81.
For the experiment in Table 5 of the main paper, we limited the option space for four options following \cite{star}. 
Notably, we observed that this reduction in the option space helped the VILA model achieve a significant performance improvement
of 38.6\% 
as shown in Table \ref{vila_options}, probably because a smaller option space reduces ambiguity and simplifies the decision-making process for the model, allowing it to focus more effectively on relevant distinctions between the choices.

\begin{table*}
  \centering
  \scriptsize
  \resizebox{1\textwidth}{!}{
  \begin{tabular}{|l|l|}
    \hline
    SSG classification task & Question \\
    \hline

    (1) Verb predicate & \textless image\textgreater What is the \{verb predicate\} between person and \{object\}? 0. option\_1 1. option\_2 ...\\
    (2) SRV $|$ object SRV & \textless image\textgreater What is the \{role\} of the \{object\}? 0. option\_1 1. option\_2 ... \\
    (3) SRV $|$ person SRV & \textless image\textgreater What is the \{role\} of the person? 0. option\_1 1. option\_2 ... \\
    (4) SRV $|$ verb predicate SRV & \textless image\textgreater What is the \{role\} used by the person for \{predicted verb predicate\} the \{object\}? 0. option\_1 1. option\_2 ...\\

    \hline
  \end{tabular}%
  }

  \caption{Various types of question prompts passed to VILA \cite{lin2024vila}, each formatted using the Vicuna v1 prompt style. For instance, the final prompt for a question appears as: \emph{"A chat between a curious user and an artificial intelligence assistant. The assistant gives helpful, detailed, and polite answers to the user's questions. USER \textless question\textgreater. Answer with the option's number from the given choices directly. ASSISTANT"}. 
  The template of questions in (4) may vary depending on the verb predicates and their semantic roles. 
  }
  \label{tab:vila_prompts}
\end{table*}

  \begin{table}
  \centering
  \scriptsize
  \resizebox{0.75\textwidth}{!}{
  \begin{tabular}{|l|c|c|}
    \hline
    SSG task & \# Training questions & \# Testing questions\\
    \hline
    
    Verb predicate classification & 39 704 & 12 123 \\ 
    Verb predicate SRV classification & 125 830 & 38 437 \\
    Object semantic SRV classification & 206 775 & 64 042 \\
    Person semantic SRV classification & 102 910 & 25 285 \\
    Total & 475 219 & 139 887 \\

    \hline
  \end{tabular}%
  }
  
  \caption{Statistics about the VQA dataset we created from our SSG dataset for evaluating VILA \cite{lin2024vila} on the SSG task.}
  \label{tab:SSG-VQA-subset}
\end{table}

\begin{table}[t!]
\centering
\resizebox{0.8\textwidth}{!}{%
\begin{tabular}{|l|c|c|c|c|}
\hline

{\shortstack{Input  }}  & {\shortstack{Obj SRV  \\ questions}}  & {\shortstack{Verb predicate \\ questions}}  & {\shortstack{Verb predicate \\ SRV questions}}  & {\shortstack{Person SRV \\ questions}}  \\ \hline

Image + question with all options & 11.3 & 19.6 & 10.9 & 34.5 \\ 
Image + question with four options & 53.9 & 60.1 & 70.6 & 46.2 \\ 

\hline
\end{tabular}%
  }
\caption{Prompting VILA \cite{lin2024vila} with MCQs having different option spaces.}
\label{vila_options}
\end{table}

\subsection{Fine-tuning CLIP ViT-L-14-336 model on SSG dataset}
\label{sec:fine-tuning clip}

When fine-tuning the CLIP ViT-L/14-336 \cite{clip} model on our SSG dataset, we generated detailed descriptions for each video frame using the SSG annotations. 
We used a single 49GB NVIDIA RTX A6000 node with four GPUs with batch size 16 and learning rate 5e-6.
We then used this model checkpoint to extract the required image and text embeddings for the InComNet-336-FT model.


\subsection{Detailed settings and further experiments on InComNet}
\label{sec:incomnet_settings}

\paragraph{Performance of InComNet across iterations:}
\label{sec:iterations}
We report InComNet's performance across iterations for object SRV classification, verb predicate classification and verb predicate SRV classification in Table \ref{tab:iteration_results}.
We first conducted iteration 1-5 by increasing the number of iterations by a factor of 1. However, as most of the metrics exhibited a fluctuating trend, we decided to increase the number of iterations by a factor of 5. Subsequently, we observed that, despite the fluctuating behaviour, the model reaches its highest overall performance with refinement of its predictions around iterations 10 and therefore, more iteration number is unnecessary \cite{fang2021read}. 
A plausible reason for the fluctuating behaviour of certain metrics could be attributed to the propagation of errors from the complementary exchange of input-output information between the sub tasks.
We applied iterations during testing as well, as it produced better results \cite{fang2021read}.
However, it should also be noted that the inference time increases as the iteration number increases.

\begin{table*}[t]
  \begin{center}
  \resizebox{1\textwidth}{!}{
  \begin{tabular}{|l|c|c|c|c|c|c|c|c|c|c|}
    \hline
    \multirow{2}{*}{Method}
    & \multicolumn{4}{c|}{Object SRV classification} &  \multicolumn{1}{c|}{Verb predicate classification} &  \multicolumn{4}{c|}{Verb predicate SRV classification} & \multirow{2}{*}{Inference time (s)}\\
    \cline{2-5} \cline{7-10}

    & \multicolumn{1}{c|}{Role-based acc.} & \multicolumn{1}{c|}{Value} &  \multicolumn{1}{c|}{Value-two} &  \multicolumn{1}{c|}{Value-all} & \multicolumn{1}{c|}{Acc.} & \multicolumn{1}{c|}{Role-based acc.} & \multicolumn{1}{c|}{Value} &  \multicolumn{1}{c|}{Value-two} &  \multicolumn{1}{c|}{Value-all}  &\\
    
    \hline

    Individual   & 45.5 &	86.2 &	54.0 &	5.7	& 66.9	& 26.9	& 65.3	& 63.1	& 23.4 & 0.96\\
    Iteration 5  & 46.2	& 85.7	& 54.7	& 5.4	& 67.0	&  28.4	& 65.8	& 64.3	& 31.4 & 1.01\\
    Iteration 10 & \underline{46.8}	& \underline{86.6}	& \textbf{55.0}	& \underline{5.9}	& \textbf{67.8}	&  \textbf{29.4}	& \textbf{66.2}	& \textbf{64.7}	& 32.0 & 1.07\\
    Iteration 15 & \textbf{47.1}	& \textbf{88.1}	& \textbf{55.0}	& \textbf{6.0}	& \underline{67.1}	& \underline{27.0}	& \underline{65.9}	& \underline{64.3}	& \textbf{32.1} & 1.18\\
    
  \hline
  \end{tabular}%
  }
  \end{center}
  \caption{Performance of InComNet across iterations. SRV refers to semantic role-values. The bold and
underlined font show the best and the second best result respectively.}
  \label{tab:iteration_results}
\end{table*}

\paragraph{Ablating the performance of InComNet's individual stages:}

\begin{table*}

  \centering
  \begin{center}
  \resizebox{1\textwidth}{!}{
  \begin{tabular}{|l|c|c|c|c|c|c|c|c|c|c|c|c|c|}
    \hline
    \multirow{2}{*}{Method} 
    & \multicolumn{4}{c|}{Object SRV classification} 
    & \multirow{2}{*}{\shortstack{Verb predicate \\ classification acc.}}
    & \multicolumn{4}{c|}{Verb predicate SRV classification} 
    & \multicolumn{4}{c|}{Person SRV classification}\\

     \cline{2-5}  \cline{7-14} 

    & \multicolumn{1}{c|}{Role-based acc.} & \multicolumn{1}{c|}{Value} &  \multicolumn{1}{c|}{Value-two} & \multicolumn{1}{c|}{Value-all} & & \multicolumn{1}{c|}{Role-based acc.} & \multicolumn{1}{c|}{Value} & \multicolumn{1}{c|}{Value-two} & \multicolumn{1}{c|}{Value-all} & \multicolumn{1}{c|}{Role-based acc.} & \multicolumn{1}{c|}{Value} & \multicolumn{1}{c|}{Value-two} & \multicolumn{1}{c|}{Value-all}\\

    \hline

(1) Obj SR: frame + obj name + role names 	& 41.5 & 86.0	& 48.4	& 5.0 & & & & & & & & &\\											
(2) Obj SR: visual prompt + obj name + role names		& 45.5	& 86.2	& 54.0	& 5.7 & & & & & & & & &\\														
(3) Verb: frame	&  & & &				& 40.2	& & & & & & & & 	\\							
(4) Verb: visual prompt	&  & & &				& 43.6	& & & & & & & &  \\								
(5) Verb: visual prompt + obj name	 & & & &					& 66.9	& & & & & & & &  \\	 								
(6) Verb SR: frame + verb Q + role names	& & & &	 & &  26.8 & 65.3 & 63.1	& 23.4   & & & & \\								
(7) Verb SR: visual prompt + verb Q + role names 	& & & &	 & &  26.9	& 65.3	& 63.4	& 28.3   & & & &\\

\hline	
InComNet-224	& 46.8	& 86.6	& 55.0	& 5.9	& 67.8	& 29.4	& 66.2	& 64.7	& 32.0	& 39.8	& 97.4	& 77.4	& 1.6 \\
InComNet-336-Frozen	& \underline{47.4}	& \underline{87.6}	& \underline{58.3}	& \underline{6.8}	& \underline{68.5}	& \underline{36.3}	& \underline{66.7}	& \underline{64.8}	& \underline{32.6}	& 42.2	& \underline{98.0}	& \underline{77.7}	& \underline{1.8}\\
InComNet-336-FT	& \textbf{49.6}	& \textbf{89.5}	& \textbf{62.1}	& \textbf{7.8}	& \textbf{70.8}	& \textbf{38.4}	& \textbf{69.0}	& \textbf{67.4}	& \textbf{38.7}	& \textbf{49.3}	& \textbf{98.5}	& \textbf{85.8}	& \textbf{1.9} \\

  \hline
  \end{tabular}%
  }
  \end{center}
  \caption{Performance of InComNet for SSG generation task on SSG dataset. The bold and underlined font show the best and the second best result respectively.}
  \label{tab:incomnet_individual_components}
\end{table*}%

In Table \ref{tab:incomnet_individual_components} we present ablations (1)-(7), where we independently train the stages of InComNet, including object SRV, verb, and verb SRV classifications, using various combinations of input features.
Ablations (1), (2), (3), (4), (6) and (7) indicate that, performance of object SRV, verb and verb SRV improves by average 2\%
when the translucent background prompt is utilized. 
Additionally, ablation (4) and (5) indicate that, performance of verb predicate improves by 23\% when the associated object name is utilized for verb predicate classification.

\paragraph{Visualizations of InComNet's Situational Scene Graph predictions:}

\begin{figure}[htbp]
    \centering
    \includegraphics[width=0.7\textwidth]{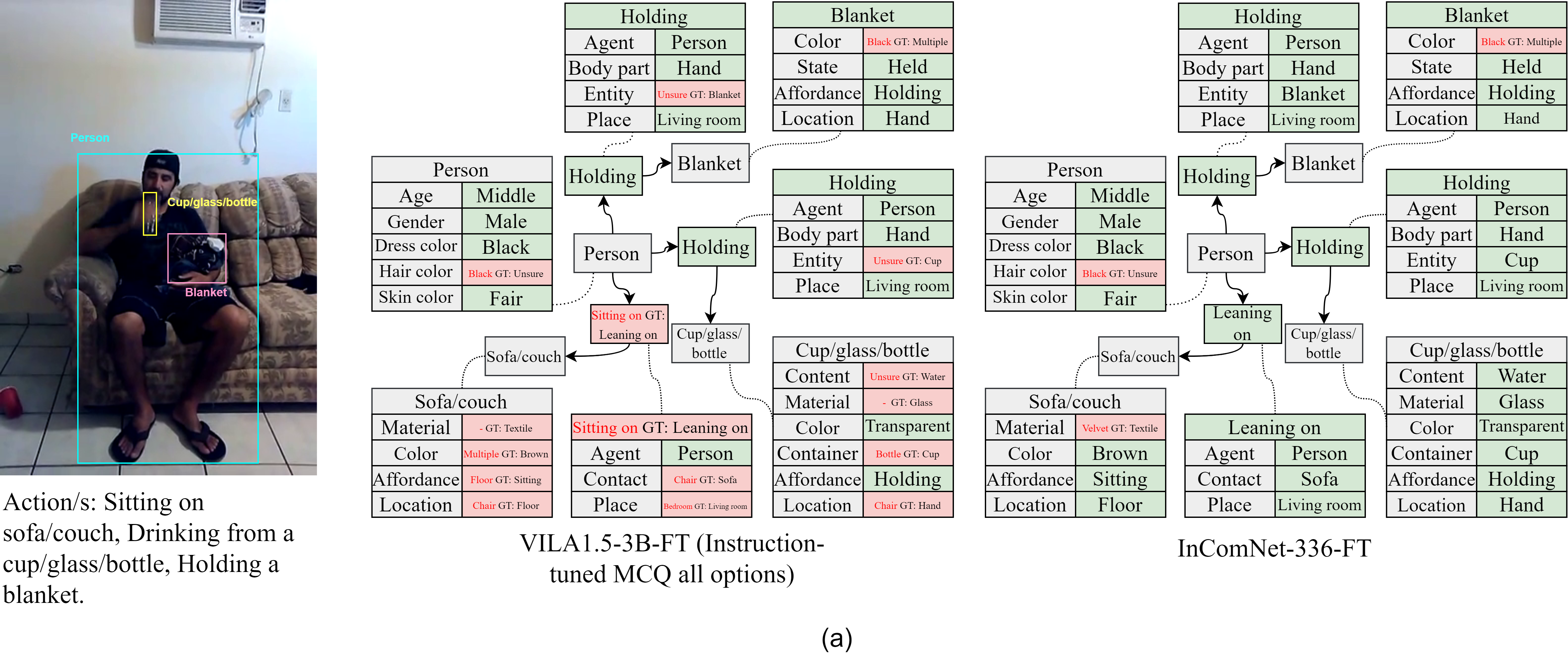}
    \vfill
    \includegraphics[width=0.7\textwidth]{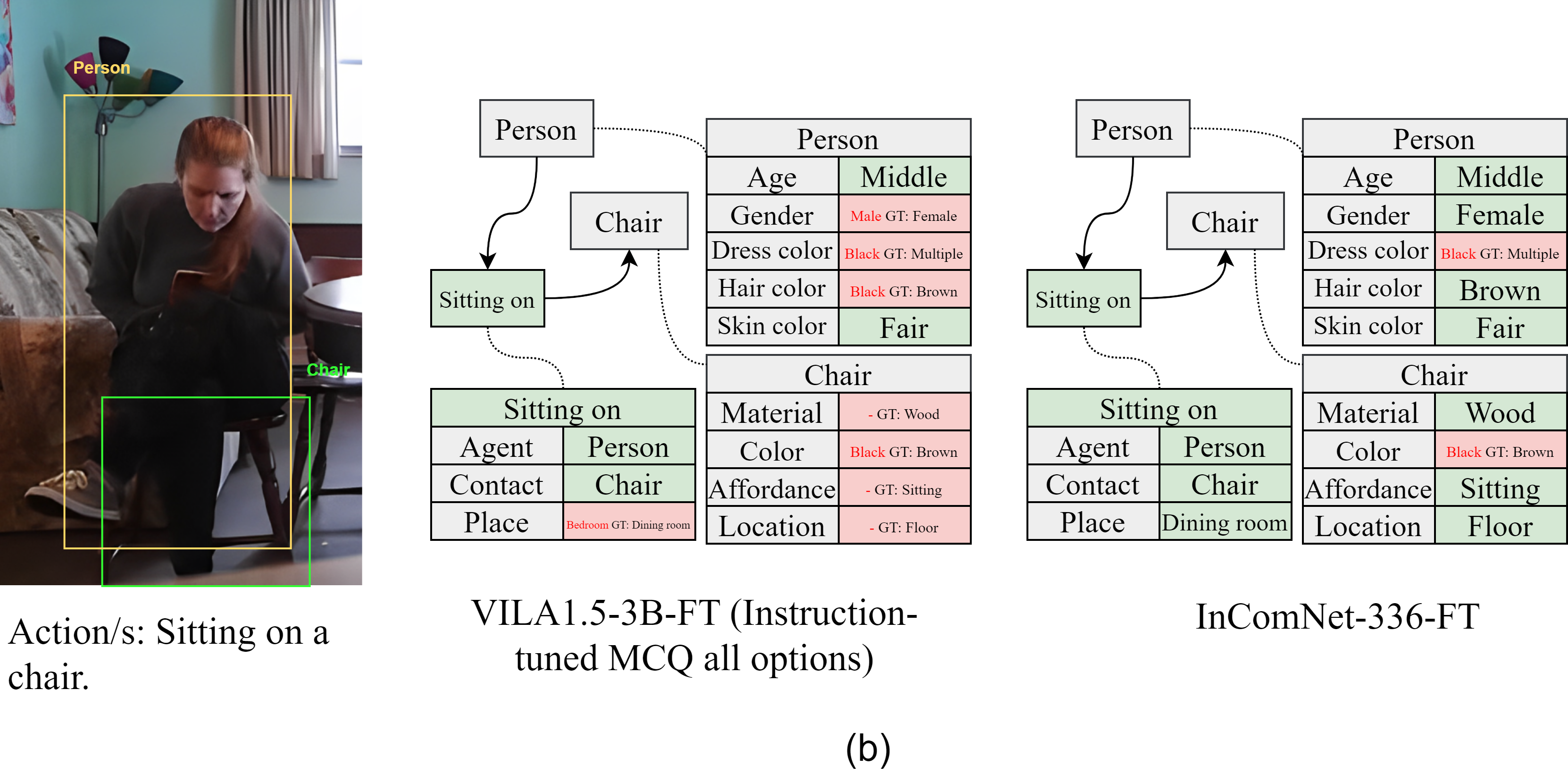}
     \vfill
    \includegraphics[width=0.7\textwidth]{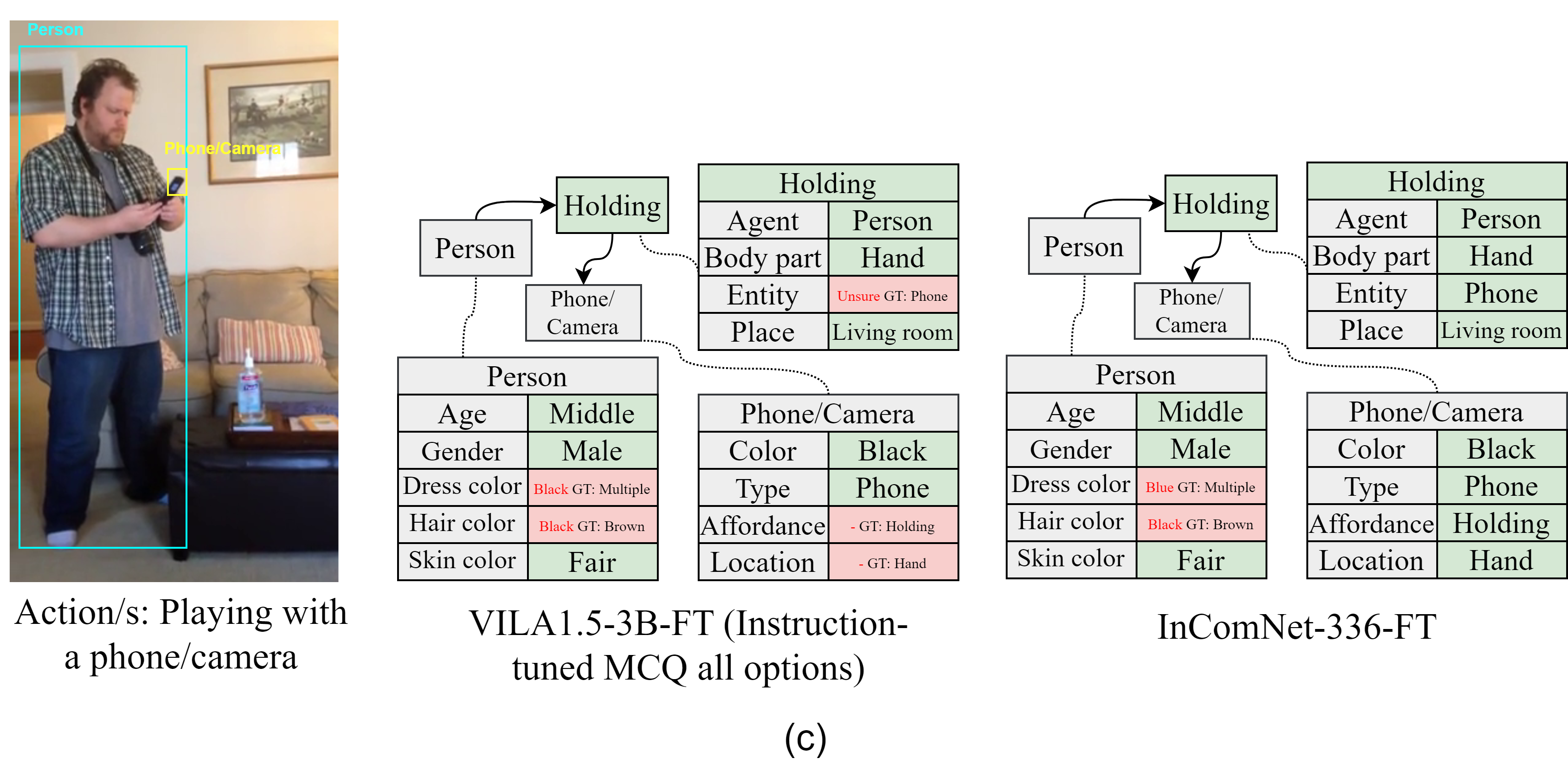}
    
      \caption{InComNet vs VILA1.5-3B-FT (Instruction-tuned MCQ all options) \cite{lin2024vila} predicted situational scene graphs. Boxes with green and red colors depict the correct and the incorrect predictions respectively.}
  \label{fig:ssg_predictions}
\end{figure}

In Fig. \ref{fig:ssg_predictions}, we compare the situational scene graph predictions of InComNet-336-FT model with the second best baseline VILA1.5-3B-FT (Instruction-tuned MCQ all options) \cite{lin2024vila} model on the situational scene graph generation task.
Overall, InComNet outperforms VILA1.5-3B-FT, particularly in predicting verb predicates and the SRV of objects, persons, and verb predicates, though minor errors may still occur. For instance, InComNet consistently predicts the correct verb predicates and their SRVs, while VILA1.5-3B-FT struggles. In Fig. \ref{fig:ssg_predictions}a, VILA1.5-3B-FT fails to predict the verb predicate correctly while it also fails to recognize certain semantic roles such as \emph{place} and \emph{entity}.
In object SRV, InComNet correctly identifies the \emph{location} precisely where as VILA1.5-3B-FT fails to do so in many cases. Further, InComNet correctly identifies the functional properties such as \emph{affordance} in all the cases where VILA1.5-3B-FT fails in most of the cases.
However, there could be errors in some predictions of InComNet.
For an example, in Fig. \ref{fig:ssg_predictions}b, InComNet-336-FT incorrectly identifies the color of the chair as \emph{black}, although it is challenging for humans as well to distinguish between \emph{black} and \emph{brown} in this context. Further, it seems the model tends to recognize one major visible color of the object when there are multiple colors e.g. in Fig. \ref{fig:ssg_predictions}a, InComNet has identified color of the blanket as \emph{black} where the ground truth is \emph{multiple}.
In person SRV classification, InComNet-336-FT correctly identifies the \emph{age, gender} and \emph{skin color} in most of the cases while it fails to recognize the \emph{dress color} and \emph{hair color} correctly. Same as in object SRV classification, it seems the model tends to identify one major visible color of the dress when there are multiple colors e.g. in Fig. \ref{fig:ssg_predictions}c, model has identified the \emph{dress color} as \emph{blue} where the ground truth is \emph{multiple}.

\paragraph{Per-class and per-role performance of InComNet on Situational Scene Graph generation task:}

In verb predicate classification, the verb predicates, \emph{\{holding, sitting on, standing on, drinking from\}} obtained the top-in-class accuracies.
In object SRV classification, the semantic roles \{\emph{material, goal, location, content}\} received the highest per-role performance while the roles \{\emph{agent, contact, body part, entity}\} obtained the highest per-role performance in verb SRV classification. In person SRV classification, the roles \{\emph{gender, skin color}\} have received the highest per-role performance.


\subsection{Hyper-parameters of InComNet}
\label{sec:hyperparamters}

We employ 4 heads and 4 layers in Transformer encoders in object SRV, verb predicate, verb predicate SRV and person SVR encoders. We also explore the effect of fixed vs learned positional embeddings and find that fixed positional embeddings give better results compared to learned positional embeddings.
As shown in Section \ref{sec:iterations}, we choose number of iterations of InComNet to be 10 as it gave the highest overall performance.
Further, InComNet-224 model has 7.5M paramaters while, InComNet-L-14-336-FT Frozen and InComNet-L-14-336-FT models have 16.6M and 449M parameters, respectively. The larger parameter count in InComNet-L-14-336-FT is due to the fine-tuning of the CLIP ViT-L-14-336 model on our SSG dataset.


\subsection{Evaluation of different visual prompts for Situational Scene Graph generation task}
\label{sec:Evaluation of different visual prompts}

\begin{figure}[htbp]
    \centering
    \includegraphics[width=0.5\textwidth]{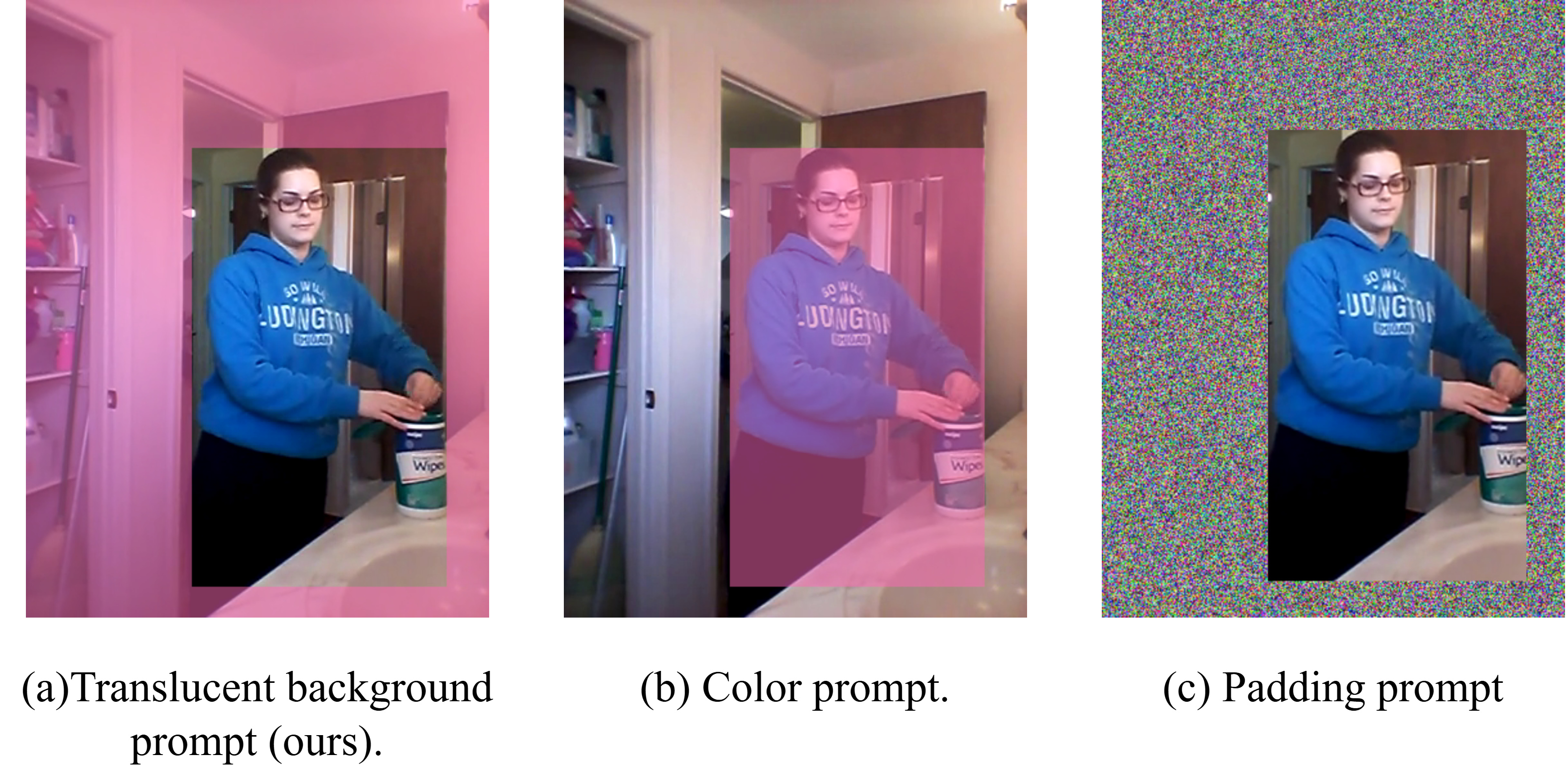}

      \caption{Different visual prompts.}
    \label{fig:prompts}
\end{figure}

In this work, we experimented the applicability of different visual prompts for situational scene graph generation task. 
In our translucent background prompt (see Fig. \ref{fig:prompts}a), we applied a translucent pink overlay to the non-interested regions in the video frame,
allowing the model to pay more attention to the interested region. 
Here the interested region refers to the bounding box region (e.g. in object SRV classification, the prompt is applied to the non-bounding box region of the object while retaining the bounding box region in its original state).
In the color prompt \cite{zhang2023fine} (see Fig. \ref{fig:prompts}b), we applied a translucent pink overlay to the interested regions in the video frame (e.g. in verb predicate classification, the prompt is applied to the union region of the bounding boxes of the person and the object). 
Following \cite{bahng2022exploring}, in the padding prompt (see Fig. \ref{fig:prompts}c) we applied a zero padding to the non-interested region in the video frame. 
As shown in Table \ref{ablation_visual_prompts}, translucent background prompt has achieved the overall best performance over the existing visual prompts.
On one hand, a plausible explanation for this behaviour could be that the translucent background introduces a certain degree of distortion to the information in the non-interested region while retaining the complete features of the interested region. This enables the model to concentrate its attention effectively on the specified region.
Specially in the verb and the verb SRV classifications, translucent background prompt likely enhances the human pose estimation due to the retention of complete features of the relationship region.
On the other hand, rather than entirely disregarding the non-interested region as seen in padding prompt, partially retaining the non-interested region may also help the model grasp broader semantics by integrating the dispersed individual semantics through out the video frame.

\begin{table}[t]
  \begin{center}
\resizebox{0.9\textwidth}{!}{
  \begin{tabular}{|l|c|c|c|c|c|c|c|c|}
  
    \hline
    Prompt & Task & Role-based accuracy & Value & Value-two & Value-all & Accuracy & mR & F1\\
    
    \hline
    
    \multirow{4}{2cm}{Color \cite{zhang2023fine}} & T1 & 39.9 & 85.9 & \underline{54.8} & \textbf{6.1}  & -    & -    & -    \\
    & T2 & -    & -    & -    & -    & \underline{67.6} & \underline{36.5} & \underline{37.8} \\
    & T3 & \underline{24.9} & 66.0 & 63.2 & 24.2 & -    & -    & -    \\
    & T4 & 33.4 & 96.7 & \underline{74.0} & 0.0  & -    & -    & -    \\

     \hline

    \multirow{4}{2cm}{Padding \cite{bahng2022exploring}} & T1 & \underline{42.0} & \underline{86.5} & 53.6 & 5.5  & -    & -    & -  \\
    & T2 & -    & -    & -    & -    & \textbf{67.8} & 30.5 & 31.2 \\
    & T3 & 24.5 & \underline{66.1} & \underline{64.2} & \underline{24.5} & -    & -    & -    \\
    & T4 & \underline{34.4} & \textbf{97.4} & 68.5 & 0.0  & -    & -    & -    \\

     \hline

    \multirow{4}{2cm}{Translucent background} & T1 & \textbf{46.8} & \textbf{86.6} & \textbf{55.0} & \underline{5.9}  & -    & -    & -    \\
    & T2 & -    & -    & -    & -    & \textbf{67.8} & \textbf{36.6} & \textbf{38.0} \\
    & T3 & \textbf{29.4} & \textbf{66.2} & \textbf{64.7} & \textbf{32.0} & -    & -    & -    \\
    & T4 & \textbf{39.8} & \textbf{97.4} & \textbf{77.4} & \textbf{1.6}  & -    & -    & -    \\

    \hline
  \end{tabular}%
  }
        
  \end{center}

  \caption{Comparison of different visual prompt engineering methods with CLIP embeddings for situational scene graph generation. T1, T2, T3 and T4 refers to object SRV, verb predicate, verb predicate SRV and person SRV classifications respectively. The bold and
underlined font shows the best and the second best result respectively.}

  \label{ablation_visual_prompts}
\end{table}


\subsection{Predicate classification with 8\% Action Genome training data}
\label{sec:pred_cls}

In Table 4 of the main paper we demonstrated how our model receives competitive performance with SOTA predicate classification models on Action genome dataset with just 8\% of Action Genome training data for which we have provided SSG annotations.
In Table \ref{tab:sg_performance_8_percent_AG_data}, we compare our method with SOTA baselines (with codes publicly available) trained on the same 8\% of Action Genome data and tested on the entire test set. 
The results showcase that our approach, which leverages situational scene graphs, still outperforms the SOTA TR$^{2}$ by 3.5\%, highlighting the strength of our representation in predicate classification.

\begin{table}[t!]
  \begin{center}
  \resizebox{0.4\textwidth}{!}{%
  \begin{tabular}{|l|c|c|c|}

    \hline
    Method &R@10 & R@20 & R@50  \\
    \hline
    TEMPURA \cite{nag2023unbiased} & 61.7 & 64.1 & 64.1 \\
    DSG-DETR \cite{feng2023exploiting} & 61.6 & 64.4 & 64.5 \\
    STTran \cite{sttran} &  63.8 & 66.6 & 66.6\\
    
    TR$^{2}$ \cite{wang2023cross} & \underline{67.3} &  \underline{70.1} &  \underline{70.1} \\
    
    \hline
    
    InComNet-336-FT & \textbf{69.4} & \textbf{72.7} & \textbf{72.7} \\
    
    \hline
  \end{tabular}%
  }
\end{center}
    \caption{Performance of SOTA predicate classification models trained on \textbf{8\%} Action Genome train set (for the frames we provided the SSG annotations) and tested on the entire Action genome test set. The bold and
underlined font show the best and the second best result respectively.}
    \label{tab:sg_performance_8_percent_AG_data}
\end{table}%


\subsection{Situational Scene Graphs for action recognition}
\label{sec:action_recognition}

We demonstrate the utility of the proposed SSG representation for action recognition on Charades \cite{charades}. This is a multi-label classification task where each video can have multiple action labels. 
We evaluate this problem under two settings as ground truth setting \cite{ag, radevski122021revisiting} and predicted setting.
In the ground truth setting we utilize all the frames in Action Genome dataset along with the frames for which we have provided SSG annotations. 
During training, for each video we create a multi-modal sequence of frame-level SSGs along with each frame's visual features such that the sequence for one frame is created by converting it into a CLIP visual token and all SSG elements into CLIP textual tokens with each token having a dimension of 512. 
This sequence forms the keys and values while a learnable query vector act as the query in a multi-headed transformer encoder. The learned query embedding is then passed through a linear classifier to classify the actions.
For the frames which do not have SSG annotations, we utilize the person, verb predicate and object in all relation instances.
In the predicted setting, we train the model using the SSG dataset and tested using the entire Action Genome dataset. During training, the keys, values and queries are formed in the same way as in the ground truth setting. During testing, we utilize the trained InComNet and person SRV models to obtain the learned query embeddings for verb predicates, object SRV, verb predicate SRV and person SRV respectively to form the key/value sequence. 
In Table \ref{tab:ar_ablation}, we ablate the contribution of different components in the situational scene graph for action recognition under ground truth setting. 
As shown in ablations (2) and (3), performance improves by 0.8\% when the components of the verb predicate relationships are further elaborated with semantic roles and values.
Further, (1) and (3) suggest that, action recognition performance improves when situational scene graph text is utilized instead of spatio-temporal scene graph text.
However, it is worth to mention that, performance of our model is not an upper bound on performance since we use ground truth SSG annotations for only 8\% Action Genome's spatio-temporal scene graph annotated frames.
In the predicted setting, our model received 54.4 \% performance.
We further compare the performance of our models with existing action recognition models on Charades dataset in Table \ref{tab:ar_sota}. 
Accordingly, it can be concluded that, structured action representation provided by the situational scene graphs seems to have a potential to capture finer details and associations in a human-centric situation leading towards improved action recognition performance. 

\begin{table}[t!]
  \begin{center}
  \resizebox{0.7\textwidth}{!}{%

          \begin{tabular}{|l|c|c|}
        \hline
                Ablation & Modality & mAP (\%)\\
            \hline
                (1) Spatio-temporal scene graph \cite{ag} & Text  & 61.9\\
                (2) Person + verb predicate + object & Text & 61.9\\
                (3) Situational scene graph & Text & 62.7\\
                (4) Frame + Spatio-temporal scene graph & Text, visual &  62.1  \\
                Our GT version: Frame + Situational scene graph & Text, visual & \textbf{62.9} \\
                \hline
        \end{tabular}%
  }
\end{center}
    \caption{Contribution of different components of situational scene graphs for action recognition.}
    \label{tab:ar_ablation}
\end{table}%

\begin{table}[t!]
  \begin{center}
  \resizebox{0.6\textwidth}{!}{%

          \begin{tabular}{|l|c|c|}
            \hline
                Method  & GFLOPS & mAP(\%) \\
                \hline
                STRG~\cite{wang2018videos} & 630	& 39.7 \\
                Timeception~\cite{hussein2019timeception} & - & 41.1 \\
                SlowFast~\cite{feichtenhofer2019slowfast} & 213 & 42.1 \\
                MULE~\cite{zhao2023open} & - & 47.2 \\
                AssembleNet++~\cite{girdhar2017actionvlad} & - & 59.9 \\
                GT version in~\cite{ag} & - & 60.3 \\
                Predicted version in~\cite{ag} & - & 44.3 \\
                MoViNet~\cite{kondratyuk2021movinets} & 100 & 63.2 \\
                \hline
                Our predicted version using InComNet-224 & 0.04 & 54.2 \\
                Our GT version & 0.04 & 62.9 \\
                \hline
        \end{tabular}%
  }
\end{center}
    \caption{SSG's action recognition performance on Charades dataset.}
    \label{tab:ar_sota}
\end{table}%

\bibliographystyle{plain}
\bibliography{egbib}
\end{document}